\documentclass[numbers,10pt]{arxiv}

\usepackage[utf8]{inputenc}
\usepackage{tabularx}
\usepackage{nicefrac}
\usepackage{graphicx, subcaption}
\usepackage{enumitem}
\usepackage{array, tabularx}
\newcolumntype{L}{>{\raggedright\arraybackslash}X}
\usepackage{fvextra}
\usepackage[most]{tcolorbox}
\usepackage{xspace}
\usepackage{bxcoloremoji}

\usepackage{graphicx}
\usepackage{enumitem}
\usepackage{caption}
\usepackage{xspace}
\usepackage{fvextra}
\usepackage{cleveref}
\usepackage{float} 
\usepackage{subcaption}
\usepackage[most]{tcolorbox}
\usepackage{booktabs}
\usepackage{adjustbox}
\usepackage{multirow}
\usepackage{siunitx}

\makeatletter
\DeclareRobustCommand\onedot{\futurelet\@let@token\@onedot}
\def\@onedot{\ifx\@let@token.\else.\null\fi\xspace}

\makeatother

\renewcommand{\paragraph}[1]{\textbf{#1}}

\usepackage{xcolor}
\usepackage{amsmath}
\usepackage{wrapfig}

\definecolor{promptblue}{RGB}{25, 84, 166}
\definecolor{thinkgreen}{RGB}{34, 139, 34}
\definecolor{thinktag}{RGB}{110, 65, 30}
\definecolor{thinkbg}{RGB}{252, 248, 243}
\definecolor{boxbg}{RGB}{252, 252, 252}

\tcbset{
    mainbox/.style={
        colback=boxbg,
        colframe=gray!40,
        arc=2mm,
        boxrule=0.8pt,
        left=12pt,
        right=12pt,
        top=12pt,
        bottom=12pt,
        fonttitle=\bfseries\large,
        coltitle=white,
        colbacktitle=gray!60
    }
}

\usepackage[T1]{fontenc}
\usepackage{amsfonts,amssymb,amsthm,mathtools}
\usepackage{needspace}

\usepackage{amsmath,amsfonts,bm}

\def\eqref#1{equation~\ref{#1}}

\def\1{\bm{1}}

\DeclareMathAlphabet{\mathsfit}{\encodingdefault}{\sfdefault}{m}{sl}
\SetMathAlphabet{\mathsfit}{bold}{\encodingdefault}{\sfdefault}{bx}{n}

\renewcommand{\paragraph}[1]{\sansbf{#1}}

\definecolor{turquoise}{RGB}{64,224,208}

\usepackage{wrapfig}
\usepackage{amsmath,amssymb}
\usepackage{tcolorbox}
\tcbuselibrary{listings,breakable,skins}
\newtcblisting{promptbox}[1]{listing only, breakable, enhanced, title={#1},
  colback=black!3, colframe=black!55, boxrule=0.4pt, arc=1.5pt,
  left=4pt, right=4pt, top=3pt, bottom=3pt, fonttitle=\bfseries\small,
  listing options={basicstyle=\scriptsize\ttfamily, breaklines=true, columns=fullflexible,
    keepspaces=true, breakindent=0pt}}
\definecolor{postleader}{HTML}{C9502E}
\definecolor{postmate}{HTML}{4D4D4D}
\newtcolorbox{forumbox}{enhanced, frame hidden, colback=black!3, boxrule=0pt, arc=0pt,
  borderline west={1.5pt}{0pt}{black!30}, left=6pt, right=4pt, top=3pt, bottom=3pt, after skip=4pt,
  before={\par\nopagebreak\addvspace{3pt}\noindent},
  before upper={\setlength{\parindent}{0pt}\setlength{\parskip}{3pt}\frenchspacing}}
\def\forumleader{agent\_2}
\newtcolorbox{taskcard}[2]{enhanced, frame hidden, colback=black!3, boxrule=0pt, arc=0pt,
  borderline west={1.5pt}{0pt}{black!30}, left=6pt, right=6pt, top=4pt, bottom=4pt, before skip=5pt, after skip=5pt,
  before upper={\setlength{\parindent}{0pt}{\sffamily\bfseries #1}\hfill{\sffamily\footnotesize\color{black!60}#2}\par\vspace{2pt}}}
\newenvironment{cardfields}{\begin{description}[style=sameline, leftmargin=5.9em, labelwidth=5.5em, labelsep=0.4em,
  font=\sffamily\footnotesize\bfseries, itemsep=1pt, parsep=0pt, topsep=1pt, partopsep=0pt]\small}{\end{description}}
\newcommand{\forumpost}[4]{\def\forumwho{#1}%
  {\footnotesize\sffamily{\bfseries\ifx\forumwho\forumleader\color{postleader}\else\color{postmate}\fi #1}%
  \enspace{\color{black!55}#2\enspace #3}}\\*{\ttfamily\footnotesize #4}\par}

\newcommand{\duke}{\ensuremath{\textcolor[HTML]{012169}{\blacklozenge}}}
\newcommand{\princeton}{\ensuremath{\textcolor[HTML]{E77500}{\clubsuit}}}
\newcommand{\nyu}{\ensuremath{\textcolor[HTML]{57068C}{\spadesuit}}}
\newcommand{\modal}{\ensuremath{\textcolor[HTML]{00A86B}{\scalebox{1.0}{$\blacktriangle$}}}}

\title{Test-Time Compute Scales Better with Parallel Agent Teams}

\author[\duke\,\modal\,*]{\href{https://openreview.net/profile?id=~Junlin_Wang1}{\textcolor{black}{Junlin Wang}}}
\author[\princeton\,*]{\href{https://openreview.net/profile?id=~Haozhe_Chen2}{\textcolor{black}{Haozhe Chen}}}
\author[\nyu\,\modal]{\href{https://openreview.net/profile?id=~Ang_Li26}{\textcolor{black}{Ang Li}}}
\author[\modal]{\href{https://openreview.net/profile?id=~Rahul_Chalamala1}{\textcolor{black}{Rahul Chalamala}}}
\author[\princeton]{\href{https://openreview.net/profile?id=~Zhuang_Liu1}{\textcolor{black}{Zhuang Liu}}}

\affiliation[\duke]{\textcolor[HTML]{012169}{Duke University}}
\affiliation[\modal]{\textcolor[HTML]{00A86B}{Modal Labs}}
\affiliation[\princeton]{\textcolor[HTML]{E77500}{Princeton University}}
\affiliation[\nyu]{\textcolor[HTML]{57068C}{New York University}}

\contribution[*]{Equal contribution}
\metadata[Code]{\url{https://github.com/modal-projects/parallel-agent-teams-scaling}}

\abstract{
Test-time scaling improves the performance of a fixed model by spending more computation at inference. There are two common ways to spend it on agents when a verifier can score every attempt. Best-of-$N$ sampling runs many independent agents and keeps the best result but saturates quickly. One large team of collaborating agents can score higher but stops improving as it grows. Can we get the best of both? Our method, \emph{parallel agent teams}, does so by sampling many small teams in parallel and keeping the best team's result. We test it on 12 long-horizon tasks from GPU kernels and formal proofs to exploit development. With a 27B model and a budget of 100 agents, parallel teams of ten beat parallel agents on 10 of 12 tasks and beat one team of 100 on all four tasks where we ran such large teams. At equal dollar cost the same parallel teams beat a frontier model on half the tasks. A simple statistical model shows that individual agents lose because their many samples have a low mean and that one large team loses because a single sample gets no gain from variance. Sampling many small teams in parallel balances a high mean with the gain from variance. Parallel agent teams thus offer a more effective way to scale test-time compute.
}

\begin{document}
\maketitle

\section{Introduction}
\label{sec:intro}

\begin{figure}[!b]
\centering
\includegraphics[width=\textwidth]{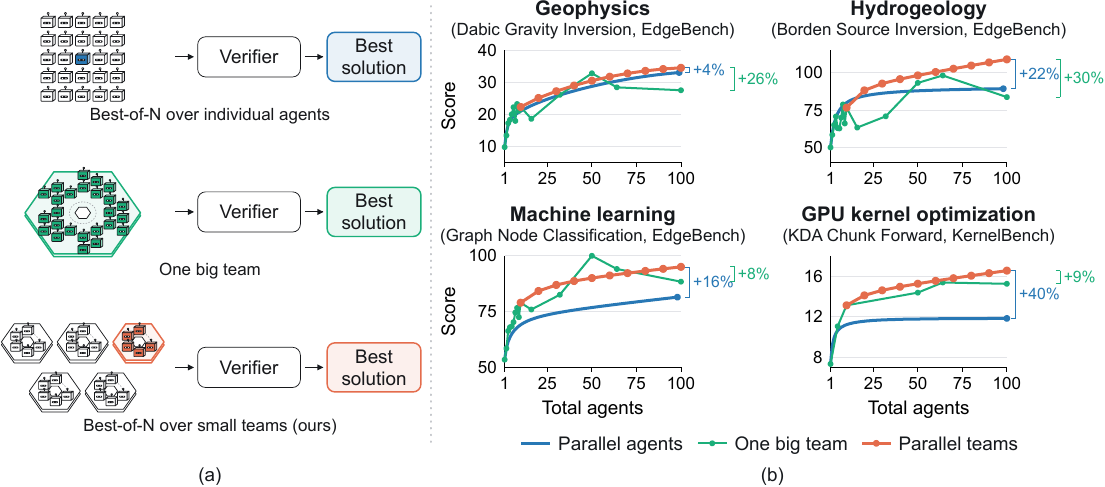}
\caption{\textbf{Sample small teams, not agents or one big team.} (a) Three ways to spend the same agents. A verifier keeps the best solution. (b) Qwen3.8-27B on four tasks. Parallel agents (blue) flatten quickly, one big team (green) rises and falls with its size, and parallel teams of ten (orange) keep improving. Brackets give the gain of parallel teams over each baseline at the largest budget both reach.}
\label{fig:headline}
\end{figure}

Test-time compute lets a fixed pre-trained model do better by spending more computation at inference \citep{snell2024scaling,wu2024inferencescaling}. An ideal method would turn more compute into better results steadily. We study tasks where a verifier can score every attempt and only the best solution matters. Many tasks for coding agents have this form \citep{chan2025mlebench,wijk2025rebench} because a scorer can check each attempt automatically and only the fastest kernel or the finished proof matters. On such tasks there are two common ways to spend more compute on agents. The first is best-of-$N$ sampling, one of the strongest methods when a verifier is available \citep{chen2021codex,cobbe2021verifiers,li2022alphacode,brown2024monkeys}. It runs $N$ independent agents and keeps the one with the highest score. Yet best-of-$N$ over individual agents saturates quickly. In our experiments, 50 agents already reach 81\% to 100\% of the gain that 100 agents reach.

The second way is to let more agents work together on the task as one team. Multi-agent systems split the work among agents with different roles \citep{hong2024metagpt,qian2024chatdev,wu2023autogen}, and recent work grows such teams to hundreds or thousands of agents \citep{qian2025macnet}. A larger team does not give more samples to select from. It changes a single sample, and its score does not rise steadily with its size \citep{qian2025macnet,kim2025scaling}. Large groups of people and of language models tend to converge on one solution and lose the diversity that exploration needs \citep{lazer2007network,wu2019largeteams,weng2025conformity}.

\begin{figure}[t]
\centering
\includegraphics[width=\textwidth]{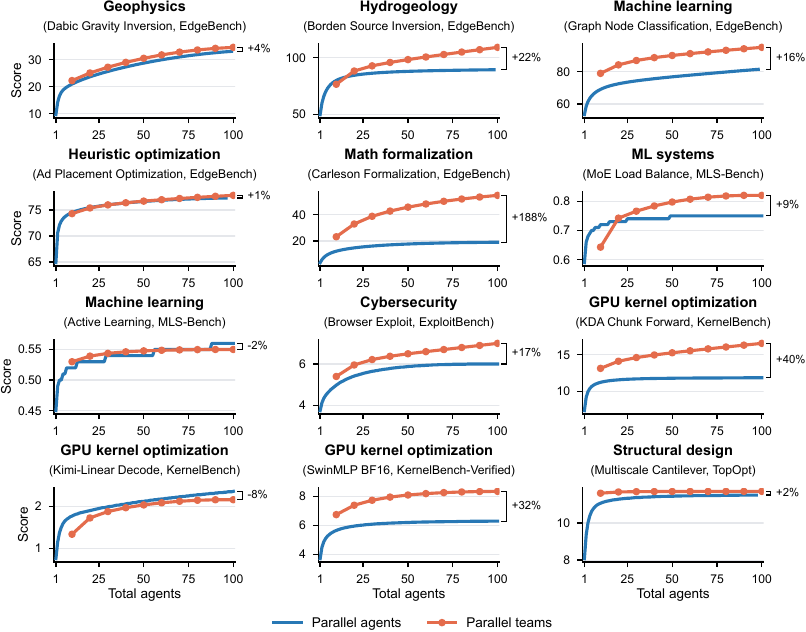}
\caption{\textbf{Parallel teams beat parallel agents on 10 of 12 tasks at 100 agents.} Qwen3.8-27B parallel agents (blue) and parallel teams of ten (orange). Brackets give the teams' gain at the largest budget both reach.}
\label{fig:teams}
\end{figure}

Each approach has one half of what we need. Best-of-$N$ takes advantage of the variation across samples. A team lets its agents work together and reach better performance more efficiently. We introduce \emph{parallel agent teams}, which combine the two by applying best-of-$N$ to independent small teams of collaborating agents (Figure~\ref{fig:headline}a). Each team is a small group of agents that share a forum. At the same budget of agents, parallel teams score higher than both parallel agents and one big team and keep improving as the budget grows (Figure~\ref{fig:headline}b). Averaged over 12 tasks, parallel teams of ten score 10\% higher than parallel agents at a total budget of 10 agents and 27\% higher at a total budget of 100 agents. At 100 agents, ten teams of ten beat one team of 100 by 8\% to 30\%. At equal dollar cost, teams of a 27B model also beat the frontier model GPT-5.6-sol on 4 of 8 tasks.

A simple statistical model of best-of-$N$ explains why parallel small teams win. The expected best of several samples is the mean score of one sample plus a gain from variance, which grows with the number of samples the budget buys. Parallel agents collect the largest gain from variance because they draw the most samples, but each agent has a low mean. One big team has a high mean, but it is a single sample and wastes the gain from variance, and a larger team does not guarantee a higher mean. Parallel small teams balance the two. Each team has a much higher mean than an agent, and several teams still collect most of the gain from variance. They also keep improving as the budget grows. At a given budget they draw fewer samples than parallel agents because each team costs several agents, so their best-of-$N$ curve is still on its steep early part, where each additional sample raises the best result the most.

Collaboration raises the mean score of each sample, and sampling turns the variance across samples into gains. Parallel agent teams get both and keep improving where best-of-$N$ over individual agents has saturated, offering a simple way to scale test-time compute further.

\section{Parallel Agent Teams}
\label{sec:setup}

We first describe best-of-$N$ sampling and how we measure it, and then agent teams. We then introduce parallel agent teams, which combine the two.

\paragraph{Best-of-\texorpdfstring{$N$}{N} sampling.} \label{sec:premise} We study tasks with a verifier that scores any candidate solution so that the best of several candidates can be found without human judgment. For a fixed model, drawing more independent responses and keeping the best steadily improves performance \citep{brown2024monkeys}. Let $X$ be the score of one sample, such as one agent, defined as the score of its best verified submission. We report the expected best of $N$ independent samples:
\begin{equation}
\mathbb{E}\Big[\max_{1\le i\le N} X_i\Big].
\label{eq:bestofn}
\end{equation}
We estimate it from a pool of $L$ runs by taking every subset of $N$ runs and averaging its best score over all $\binom{L}{N}$ subsets. With the scores sorted as $v_{(1)}\le\dots\le v_{(L)}$, this average is $\sum_{i=N}^{L} v_{(i)}\binom{i-1}{N-1}/\binom{L}{N}$. One pool thus gives the whole curve from $N=1$ to $N=L$. Take five runs that score 2, 4, 5, 7 and 9. Their expected best is 5.4 for one run (the mean), 7.1 for two and 9 for all five.

\paragraph{Agent teams.} \label{sec:setup-team} Instead of working alone, agents can work together as a team on the same task. Past work shows that such teams can produce better results than a single agent \citep{li2023camel,du2024debate,hong2024metagpt,qian2024chatdev,chen2024agentverse}. We mainly study flat teams, in which peers share progress through a common log or memory \citep{park2026communication,qu2026coral}. Each agent has its own sandbox and context but shares a forum where any agent can post and read notes at any time. There are no roles, no manager and no assigned work. A team's score is the best verified submission of any member (Appendix~\ref{app:forum}). Section~\ref{sec:ablate-harness} studies an alternative team architecture in which a manager assigns the work.

\paragraph{Parallel agent teams.} Parallel agent teams combine the two by applying best-of-$N$ to independent teams (Figure~\ref{fig:headline}a), so a sample is one team. We measure compute as a budget of $B$ agents that each run for the same fixed time. It buys $N=B/k$ teams of size $k$. Parallel agents ($k=1$) and one big team ($k=B$) are the two ends of this family, and parallel teams lie in between. We use teams of ten unless stated otherwise and study the team size in Section~\ref{sec:res-size}.

\begin{table}[t]
\caption{\textbf{The 12 tasks.} Higher is better for every metric.}
\label{tab:tasks}
\centering\footnotesize
\setlength{\tabcolsep}{4pt}
\begin{tabular}{llll}
\toprule
Task & Benchmark & Domain & Metric \\
\midrule
Dabic Gravity Inversion & EdgeBench & Geophysics & points 0--100 \\
Borden Source Inversion & EdgeBench & Hydrogeology & points 0--100 \\
Graph Node Classification & EdgeBench & Machine learning & points 0--100 \\
Ad Placement Optimization & EdgeBench & Heuristic optimization & points 0--100 \\
Carleson Formalization & EdgeBench & Math formalization & points 0--100 \\
MoE Load Balance & MLS-Bench & ML systems & reward 0--1 \\
Active Learning & MLS-Bench & Machine learning & reward 0--1 \\
Browser Exploit & ExploitBench & Cybersecurity & ladder 0--16 \\
KDA Chunk Forward & KernelBench & GPU kernel optimization & 1/latency \\
Kimi-Linear Decode & KernelBench & GPU kernel optimization & 1/latency \\
SwinMLP BF16 & KernelBench-Verified & GPU kernel optimization & speedup \\
Multiscale Cantilever & TopOpt & Structural design & $C_{\mathrm{ref}}/C$ \\
\bottomrule
\end{tabular}
\end{table}

\section{Parallel Small Teams Beat Both}
\label{sec:results}

\begin{figure}[t]
\centering
\includegraphics[width=\textwidth]{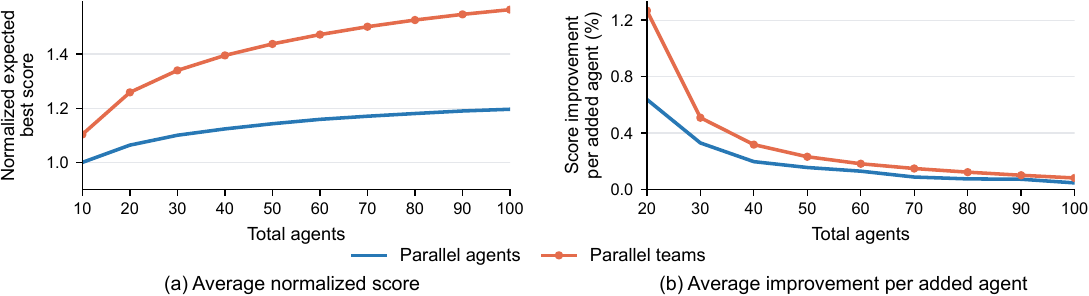}
\caption{\textbf{Averaged over 12 tasks, parallel teams score higher and keep improving for longer.} Qwen3.8-27B parallel agents (blue) and parallel teams of ten (orange). (a) Expected best score normalized by that of ten agents. The gap widens with the budget. (b) Parallel teams gain more from each added agent.}
\label{fig:gains}
\end{figure}

\paragraph{Setup.} \label{sec:setup-tasks} We use 12 long-horizon optimization tasks from six benchmarks \citep{edgebench2026,lyu2026mlsbench,lee2026exploitbench,kernelbenchcom,kernelbenchverified,ouyang2025kernelbench,chen2025multiscale}. They span nine domains from GPU kernels to formal mathematics (Table~\ref{tab:tasks}). In each task an agent gets a description, a working environment and a scorer it can call and then improves its submission for 12 hours. Each agent runs in the Codex harness \citep{openai2025codex} inside its own sandbox. Our main model is Qwen3.8-27B \citep{qwen2026qwen38}. Section~\ref{sec:other} repeats the comparison with a stronger model, DeepSeek V4.1 Flash \citep{deepseek2026v41flash}, and Section~\ref{sec:cost} compares against GPT-5.6-sol \citep{openai2026gpt56}. Each task has about 100 independent runs and 10 teams of ten, and four tasks also have single teams of 16 to 100 agents (Appendix~\ref{app:infra}).

\subsection{Better than one big team}
\label{sec:res-size}

Figure~\ref{fig:headline}b adds one team of $B$ agents on the four tasks where we ran one big team. Its score rises at first but not steadily. From 64 to 100 agents it drops on all four tasks. At 100 agents, ten teams of ten beat one team of 100 on all four tasks by 8\% to 30\%, and on two tasks one team of 100 even scores below the best of 100 independent agents.

\begin{figure}[t]
\centering
\includegraphics[width=\textwidth]{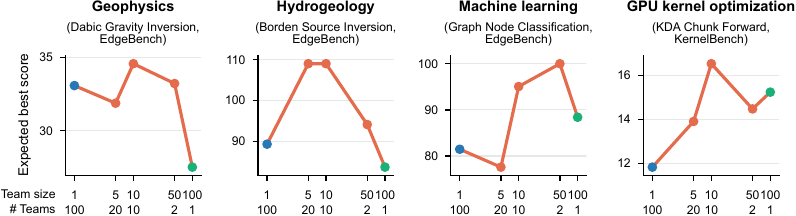}
\caption{\textbf{At a fixed budget, the best team size lies in between.} 100 Qwen3.8-27B agents form $100/k$ teams of size $k$. The ends are parallel agents ($k=1$; blue) and one big team ($k=100$; green).}
\label{fig:sweep}
\end{figure}

Figure~\ref{fig:sweep} holds the budget at 100 agents and varies the team size. The best size lies in between on every task. Ten teams of ten score highest on three tasks, and two teams of 50 on Graph Node Classification, where both reached the maximum score. Section~\ref{sec:why} explains why both ends lose.

\subsection{Better than parallel agents}
\label{sec:res-teams}

Figure~\ref{fig:teams} compares parallel agents with parallel teams of ten on each task. At a budget of 50 agents, the best of five teams beats the best of 50 agents on 11 of 12 tasks. At 100 agents, the best of ten teams beats the best of 100 agents on 10 of 12 tasks (Table~\ref{tab:k10}). The gains are largest where individual agents rarely succeed. On Carleson Formalization the best of 100 agents scores 19.0 out of 100 while the best team scores 54.8. Averaged over the 12 tasks, parallel teams score 10\% higher than parallel agents at a total budget of 10 agents and 27\% higher at 100 agents (Table~\ref{tab:k10}). Parallel agents beat parallel teams only on Kimi-Linear Decode and, by a small margin, on Active Learning.

\subsection{Slower saturation than parallel agents}
\label{sec:res-scale}

Parallel agents improve monotonically as compute grows because an extra sample can only raise the best score, but they saturate quickly. At 20 agents, each task has already reached 60\% to 94\% of the gain that about 100 agents bring. Parallel teams keep improving for longer (Figure~\ref{fig:gains}b). From 10 to 20 agents each extra agent adds 1.27\% to the score of parallel teams but only 0.64\% to the score of parallel agents. The gap between the two methods therefore widens as the budget grows.

\subsection{Beating a frontier model}
\label{sec:cost}

\begin{figure}[t]
\centering
\includegraphics[width=\textwidth]{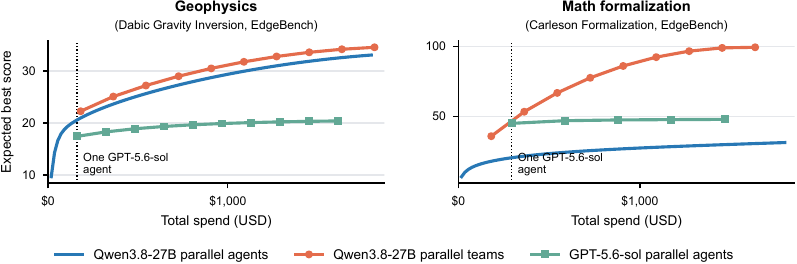}
\caption{\textbf{At equal cost, parallel teams of a 27B model overtake a frontier model.} Qwen3.8-27B parallel agents (blue) and parallel teams of ten (orange), priced by GPU time, against GPT-5.6-sol agents (green), priced by tokens. The dotted line marks the price of one GPT-5.6-sol agent.}
\label{fig:cost}
\end{figure}

We also compare teams with GPT-5.6-sol at equal dollar cost (Figure~\ref{fig:cost}). A Qwen3.8-27B team of ten costs about as much as one or two GPT-5.6-sol agents (Appendix~\ref{app:cost}). On Carleson Formalization one GPT-5.6-sol agent beats one Qwen3.8-27B team, yet for the price of five GPT-5.6-sol agents, eight Qwen3.8-27B teams reach 52.0 against 47.9. Across the eight tasks with at least five GPT-5.6-sol runs, Qwen3.8-27B teams beat five GPT-5.6-sol agents at the same price on four. GPT-5.6-sol scores high but varies little from run to run. More GPT-5.6-sol runs therefore add little while the best of several Qwen3.8-27B teams keeps climbing.

\section{Why Do Parallel Teams Win?}
\label{sec:why}

Parallel teams beat both common approaches. They beat parallel agents although at the same budget they draw ten times fewer samples, since a team of ten costs ten agents. They beat one big team, which puts the whole budget into a single sample. A simple statistical model of best-of-$N$ explains both results.

\subsection{A statistical model}
\label{sec:why-model}

Assume that the score $X$ of one sample follows a normal distribution $\mathcal{N}(\mu,\sigma^2)$ and that samples are independent. The expected best of $N$ samples is then $\mu+\sigma a_N$. Here $a_N$ is the expected maximum of $N$ standard normal variables, and $a_N\approx\sqrt{2\ln N}$ for large $N$ \citep[Appendix~\ref{app:normal};][]{fisher1928limiting,gnedenko1943limite}. Now let each sample cost $c$ agents so that a budget of $B$ agents buys $N=B/c$ samples. The expected best is then
\begin{equation}
V(B)\approx \mu+\sigma\sqrt{2\ln (B/c)}.
\label{eq:normal-budget}
\end{equation}
The three numbers affect the curve at different rates. The mean $\mu$ adds the same amount at every budget. The standard deviation $\sigma$ multiplies the gain from variance. This gain grows only as $\sqrt{2\ln N}$. The cost $c$ cuts the number of samples from $B$ to $B/c$ but lowers the gain from variance by only about $\sigma\ln c/\sqrt{2\ln B}$. A tenfold cost therefore matters little once the budget is large. For teams of size $k$ the cost is $c=k$, and the mean $\mu_k$ and standard deviation $\sigma_k$ depend on $k$:
\begin{equation}
V_k(B)\approx \mu_k+\sigma_k\sqrt{2\ln (B/k)}.
\label{eq:team-size}
\end{equation}
The first term is the mean of one team and the second is the gain from variance when sampling $B/k$ teams. Parallel agents ($k=1$) and one big team ($k=B$) are the two ends of this family.

\subsection{One big team wastes variance}
\label{sec:why-var}

Parallel sampling takes advantage of the variance of scores. By Equation~\ref{eq:normal-budget}, more samples raise the expected best by $\sigma\sqrt{2\ln N}$, so performance scales with compute at a rate set by the standard deviation $\sigma$. Parallel agents and parallel teams both draw more samples as the budget grows and collect this gain. One big team wastes it. It is a single sample, so $\sqrt{2\ln(B/k)}=0$ and its expected score is just its mean $\mu_B$, whatever its standard deviation. A larger budget then helps one big team only if a larger team has a higher mean, and a larger team does not guarantee that. From 64 to 100 agents the mean of one team falls on all four tasks where we ran such teams (Section~\ref{sec:res-size}). One big team therefore loses to parallel teams, which raise the mean and still keep most of the gain from variance.

\subsection{Parallel teams balance mean and variance}
\label{sec:why-mean}

\begin{wrapfigure}{r}{0.5\textwidth}
\vspace{-1.2em}
\centering
\includegraphics[width=0.49\textwidth]{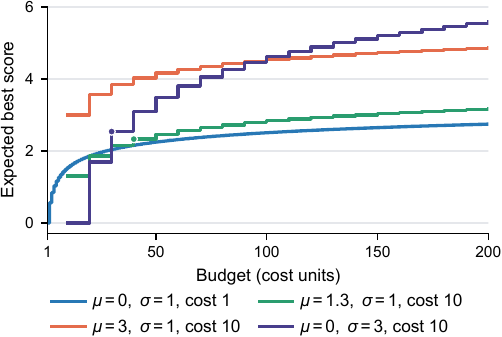}
\caption{\textbf{How the mean, standard deviation and cost of a sample shape the best-of-$N$ curve.} Expected best score in the statistical model for four sampling methods. A higher mean shifts the curve up, a larger standard deviation makes it rise faster, and a higher cost delays its start.}
\label{fig:theory}
\end{wrapfigure}

Parallel teams balance the gain from the mean and the gain from variance. Parallel agents collect the largest gain from variance, but one agent has a low mean. A team of ten raises the mean but draws fewer samples and so gives up some gain from variance. By Equation~\ref{eq:normal-budget}, an expensive method $(\mu_2,\sigma_2)$ at cost $c$ beats a cheap method $(\mu_1,\sigma_1)$ at cost 1 at budget $B$ when
\begin{equation}
\mu_2+\sigma_2\sqrt{2\ln (B/c)} > \mu_1+\sigma_1\sqrt{2\ln B}.
\label{eq:condition}
\end{equation}
With equal standard deviations, drawing $B/c$ samples instead of $B$ loses only about $\sigma\ln c/\sqrt{2\ln B}$, and this loss shrinks as the budget grows. A higher mean counts in full at every budget. At a budget of 200, a mean gain of only $0.88\sigma$ pays for a 10$\times$ cost. Figure~\ref{fig:theory} shows how the mean, standard deviation and cost each change the curve. A method at ten times the cost can still win through a higher mean or a higher standard deviation.

\subsection{Testing the statistical model}
\label{sec:why-fit}

\begin{figure}[b]
\centering
\includegraphics[width=\textwidth]{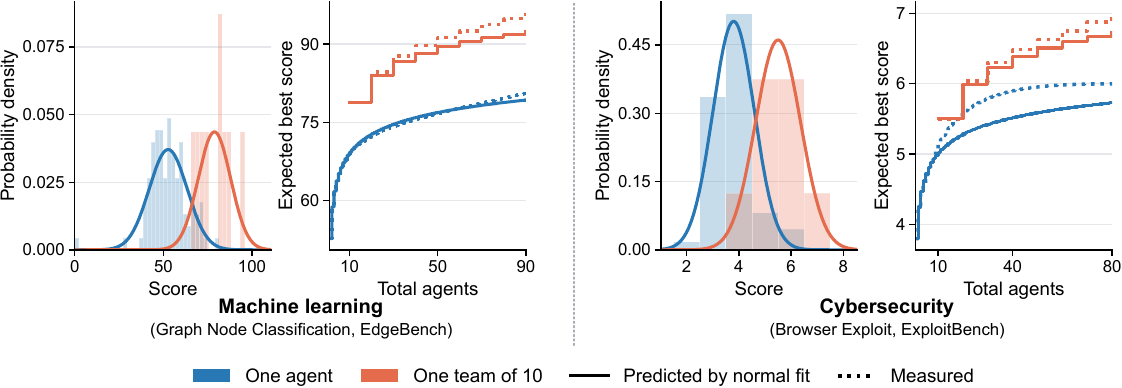}
\caption{\textbf{The statistical model predicts the measured curves when scores are close to normal.} Left of each task: score distributions of one Qwen3.8-27B agent (blue) and one team of ten (orange) with normal fits. Right: predicted (solid) and measured (dotted) expected best score.}
\label{fig:normal-fit}
\end{figure}

We now check the statistical model against our measurements. On Graph Node Classification, a team of ten scores 79.0 on average and an agent 52.1 (Table~\ref{tab:normal-gain}). With the agents' standard deviation $\sigma=11.8$, the team's mean is $2.3\sigma$ higher. Equation~\ref{eq:condition} requires only $1.4\sigma$ at a budget of 100 agents. Scores on this task and on Browser Exploit are close to normal, and the fitted normals predict the measured curves well (Figure~\ref{fig:normal-fit}).

Table~\ref{tab:normal-gain} extends the check to all 12 tasks. A team's mean exceeds an agent's by $0.5\sigma$ to $4.4\sigma$ on every task, and the statistical model predicts the sign of the teams' gain at 100 agents on 9 of 12. It overestimates the best of many agents where agent scores are far from normal, as with the long lower tail of failed runs on Ad Placement Optimization ($W=0.66$) or the pile-up below a ceiling on KDA Chunk Forward (Appendix~\ref{app:normal-all}). The statistical model also explains the frontier model's flat curve: a high mean with a small standard deviation leaves little for parallel sampling to gain.

\begin{table}[t]
\caption{\textbf{The statistical model on all 12 tasks.} Mean $\mu$ and standard deviation $\sigma$ of one Qwen3.8-27B agent and one team of ten. $W$ is the Shapiro--Wilk statistic \citep{shapiro1965normality}, which equals 1 for a normal sample. Gain is the expected best of 10 teams over that of 100 agents. Italic marks a predicted gain with the wrong sign.}
\label{tab:normal-gain}
\centering\footnotesize
\setlength{\tabcolsep}{3.5pt}
\begin{tabular}{lrrrrrrrr}
\toprule
& \multicolumn{3}{c}{One agent} & \multicolumn{3}{c}{One team of ten} & \multicolumn{2}{c}{Gain at 100 agents} \\
\cmidrule(lr){2-4} \cmidrule(lr){5-7} \cmidrule(lr){8-9}
Task & $\mu$ & $\sigma$ & $W$ & $\mu$ & $\sigma$ & $W$ & predicted & measured \\
\midrule
Dabic Gravity Inversion & 9.5 & 8.6 & 0.79 & 22.3 & 5.4 & 0.73 & \textit{-1.4\%} & +4.6\% \\
Borden Source Inversion & 49.1 & 18.9 & 0.96 & 76.5 & 19.5 & 0.93 & +10.5\% & +22.1\% \\
Graph Node Classification & 52.1 & 11.8 & 0.89 & 79.0 & 8.5 & 0.96 & +12.7\% & +17.0\% \\
Ad Placement Optimization & 62.0 & 17.9 & 0.66 & 74.3 & 1.8 & 0.99 & \textit{-28.0\%} & +0.7\% \\
Carleson Formalization & 3.5 & 4.5 & 0.78 & 23.2 & 15.7 & 0.93 & +222.7\% & +187.7\% \\
MoE Load Balance & 0.566 & 0.146 & 0.80 & 0.643 & 0.222 & 0.59 & +5.7\% & +9.2\% \\
Active Learning & 0.451 & 0.083 & 0.70 & 0.530 & 0.015 & 0.87 & -15.9\% & -1.6\% \\
Browser Exploit & 3.75 & 0.75 & 0.82 & 5.40 & 0.92 & 0.90 & +20.8\% & +16.7\% \\
KDA Chunk Forward & 7.27 & 3.27 & 0.93 & 13.12 & 1.61 & 0.96 & +0.8\% & +39.8\% \\
Kimi-Linear Decode & 0.724 & 0.712 & 0.83 & 1.334 & 0.646 & 0.92 & -7.2\% & -7.9\% \\
SwinMLP BF16 & 3.58 & 1.50 & 0.97 & 6.74 & 1.02 & 0.96 & +13.2\% & +32.5\% \\
Multiscale Cantilever & 7.81 & 2.58 & 0.91 & 11.60 & 0.12 & 0.77 & \textit{-17.5\%} & +1.7\% \\
\bottomrule
\end{tabular}

\end{table}

\subsection{Fewer samples, steeper gains}
\label{sec:why-scale}

The same statistical model explains why teams scale better. The best-of-$N$ curve rises fastest at small $N$ and flattens as $N$ grows. At a given budget an expensive method has drawn few samples and still sits on the steep part of its curve while a cheap method is already on the flat part. Differentiating Equation~\ref{eq:normal-budget} gives the return to one more agent of budget, $\sigma/\big(B\sqrt{2\ln(B/c)}\big)$. This return grows with the cost $c$. At 20 agents the statistical model gives a team of ten ($N=2$) a marginal return $\sqrt{\ln 20/\ln 2}\approx2.1$ times that of an agent ($N=20$) at equal standard deviation. The measured ratio is 1.27\% to 0.64\% (Figure~\ref{fig:gains}b). Earlier work observes a trade-off between the quality and the quantity of samples at equal compute \citep{brown2024monkeys,snell2024scaling,wu2024inferencescaling}. Our statistical model makes this trade-off precise for best-of-$N$.

\section{Other Ways to Spend Compute}
\label{sec:other}

\begin{figure}[t]
\centering
\begin{minipage}[t]{0.49\textwidth}
\centering
\includegraphics[width=\linewidth]{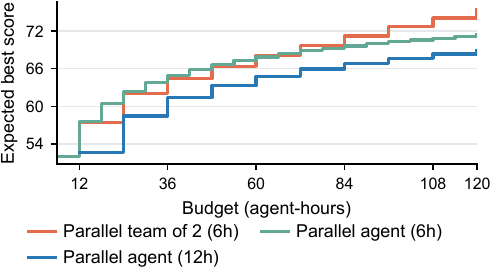}\\[-0.2em]
{\small (a) Expected best score at equal agent-hours}
\end{minipage}\hfill
\begin{minipage}[t]{0.49\textwidth}
\centering
\includegraphics[width=\linewidth]{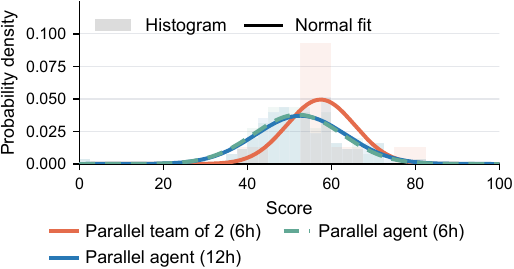}\\[-0.2em]
{\small (b) Score distributions and normal fits}
\end{minipage}
\caption{\textbf{Two agents for 6 hours beat one agent for 12 hours.} Qwen3.8-27B on Graph Node Classification. (a) Expected best score against agent-hours. (b) A longer session barely shifts the score distribution, while a team of two at the same cost shifts it to the right.}
\label{fig:duration}
\end{figure}

Equation~\ref{eq:condition} applies to parallel sampling with any way of producing a sample. We use it to evaluate two other choices, a longer session for each agent and a different model.

\paragraph{Longer sessions.} \label{sec:duration} Instead of forming a team we can let each agent work longer. A 12-hour agent costs as much as a team of two agents that each work 6 hours (Figure~\ref{fig:duration}, Appendix~\ref{app:duration}). Longer sessions barely raise the mean. On Graph Node Classification an agent's mean score rises only from 52.0 at 6 hours to 52.7 at 12, in line with the slowdown of long agent sessions \citep{liu2026slowdown}. By Equation~\ref{eq:condition}, doubling the cost pays off at this budget only if the mean rises by about 3.4 points. The 12-hour agents therefore lose even to plain best-of-$N$ over 6-hour agents. A team of two spends the same hours on two approaches, raises the mean by 5.4 points and beats both.

\begin{figure}[t]
\centering
\includegraphics[width=\textwidth]{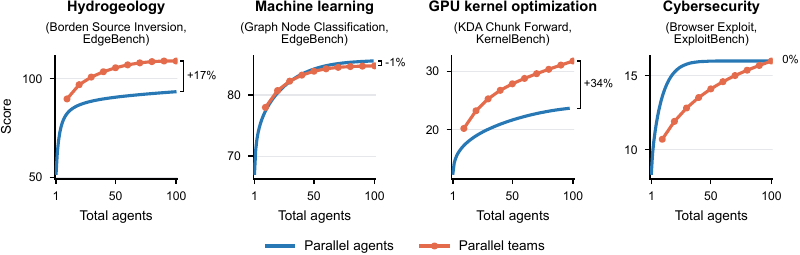}
\caption{\textbf{With DeepSeek V4.1 Flash, teams gain less on tasks that are easy for the model.} Parallel agents (blue) and parallel teams of ten (orange) on four of the eight tasks in Figure~\ref{fig:deepseek}. Brackets give the teams' gain at the largest budget.}
\label{fig:deepseek-row}
\end{figure}

\paragraph{A different model.} \label{sec:other-model} A different model changes the mean and standard deviation of every way of producing a sample. Figure~\ref{fig:deepseek-row} repeats the comparison with DeepSeek V4.1 Flash (Appendix~\ref{app:models} shows all eight tasks). A single agent already has a high mean and a small standard deviation on tasks that are easy for the model. Its best-of-$N$ curve then reaches the top score quickly and leaves a team little room to add. On tasks that are still hard for the model, a team raises the mean a lot and parallel teams beat parallel agents clearly.

\section{Inside a Team}
\label{sec:inside}

A team of ten scores higher than the best of ten independent agents. On Graph Node Classification, one team scores 79.0 on average and the best of ten independent agents scores 69.1. Eight of the 12 tasks show the same pattern (Table~\ref{tab:k10}), and so do most smaller teams (Appendix~\ref{app:bestofk}). Both use ten agents for the same time, so the extra gain must come from what agents share.

\paragraph{Sharing discoveries.} \label{sec:comm} To see what agents share, we analyzed the forum posts of our team runs (Appendix~\ref{app:conv}). Agents mostly report what they tried and the score it reached. From the first hour, they mark these reports with short tags such as \textsc{win} and \textsc{fail} that make them quick to scan (Figure~\ref{fig:tags}). Useful habits spread without assigned roles. On Borden Source Inversion 88\% of teammates start announcing experiments before running them once one agent does (Figure~\ref{fig:protocols}). Agents also build on each other's work. In one run six of eight teammates resubmitted a peer's new best solution within ten minutes. Agents agree with a named teammate more than three times as often as they disagree (Figure~\ref{fig:stance}). The same copying pulls a team toward one solution, which helps a small team but makes a large team converge (Section~\ref{sec:res-size}).

\begin{figure}[t]
\centering
\includegraphics[width=\textwidth]{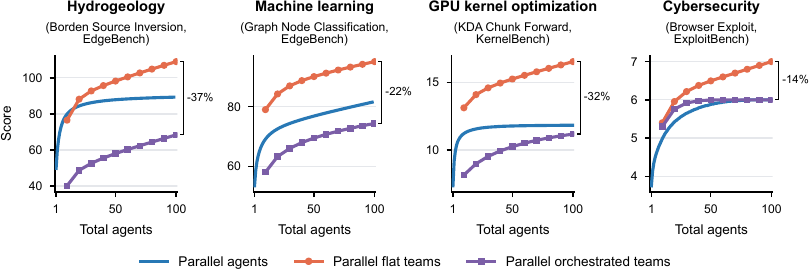}
\caption{\textbf{A manager makes teams worse.} Qwen3.8-27B parallel agents (blue), parallel flat teams of ten (orange) and parallel orchestrated teams of ten (purple), in which one agent assigns the work. Brackets give the difference from flat teams. We omit Dabic Gravity Inversion because its evaluator often crashed.}
\label{fig:orch}
\end{figure}

\paragraph{A manager hurts.} \label{sec:ablate-harness} In our flat teams no agent manages the others (Section~\ref{sec:setup}). To test this choice, we replaced the forum with an orchestrator. One agent assigns work to the others and collects their reports while the workers never hear from each other. These orchestrated teams stay below both flat teams and independent agents at every budget on Borden Source Inversion, Graph Node Classification and KDA Chunk Forward (Figure~\ref{fig:orch}). On Borden Source Inversion the best of ten orchestrated teams (68.4) scores below the average flat team (76.5). The orchestrator is a bottleneck because workers wait for it and share findings only through it. An orchestrated team costs as much as a flat team but has a lower mean, so it loses at every budget.

\section{Related Work}
\label{sec:related}

\paragraph{Scaling test-time compute.} A classic way to spend more compute at inference is to draw many samples and let a trained verifier or a majority vote pick one \citep{cobbe2021verifiers,wang2023selfconsistency,lightman2024verify,li2024moreagents}. \citet{brown2024monkeys} measured how coverage grows with the number of samples. Later work explains these curves and their limits \citep{schaeffer2025powerlaws,stroebl2026limits,li2026tailsearch}, splits a budget across models, samples and search methods \citep{snell2024scaling,wu2024inferencescaling,zhang2024osca}, and finds that without a verifier a simple majority vote beats more elaborate sequential schemes \citep{wang2025rethinking}. In best-of-$N$ the samples are independent. Tree search and evolutionary program search let later samples build on earlier ones through a fixed search procedure \citep{yao2023tot,zhou2024lats,romeraparedes2024funsearch,novikov2025alphaevolve}. Agents can now finish much longer tasks \citep{wijk2025rebench,kwa2025longtasks}, but the return per token falls within one session \citep{liu2026slowdown}.

\paragraph{Multi-agent systems.} Most multi-agent systems give agents roles and route work between them \citep{hong2024metagpt,qian2024chatdev,wu2023autogen,hu2024adas}, and fixed graphs of agents have grown past a thousand \citep{qian2025macnet}. Debate and Mixture-of-Agents combine several models on one answer \citep{du2024debate,wang2024moa}. Such systems often fail because of their design and weak verification \citep{cemri2025fail}. Debate does not reliably beat simpler prompting \citep{smit2024mad}. Evaluations that ignore cost reward needlessly complex agents \citep{kapoor2024agents}. Our teams have no roles, manager or fixed graph.

\paragraph{Team size and diversity.} Growing a team of language model agents gives diminishing and sometimes negative returns. The performance of MacNet follows a logistic curve as agents are added \citep{qian2025macnet}, and the benefit of coordination ranges from large gains to large losses across tasks \citep{kim2025scaling}. Agents conform more to a larger majority \citep{weng2025conformity}, teams settle on a compromise below their best member \citep{pappu2026experts}, and interaction shrinks the diversity of ideas \citep{chen2026diversitycollapse}. Studies of human groups find the same trade-off. Dense communication spreads good solutions quickly but ends exploration early \citep{lazer2007network,bernstein2018intermittent}. Large teams develop existing ideas while small teams find new ones \citep{wu2019largeteams}. Splitting an organization into separate subgroups keeps its diversity \citep{fang2010subgroups}. Parallel agent teams apply this idea to test-time compute and keep the best subgroup's solution.

\paragraph{Concurrent work.} \citet{park2026communication} let identical peers share verified progress through a common log. They compare one team of up to five with the best of as many independent agents. CORAL \citep{qu2026coral} runs a few agents that share a memory on open-ended optimization problems. We sample many small teams at a fixed budget, compare them with one big team and explain why they win.

\section{Conclusion and Discussion}
\label{sec:limits}

Best-of-$N$ over individual agents and one big team are the two ends of one family of ways to spend a budget of agents. Neither end scales well. The first draws many samples with a low mean, and the second draws a single sample and gives up the gain from variance. Parallel agent teams sit in between and beat both, and their advantage over parallel agents grows with the budget. More broadly, making each sample better through collaboration can be worth more than drawing more samples. A few runs of any sampling method give its mean and standard deviation, and Equation~\ref{eq:condition} then tells whether it is worth its cost. Our teams collaborate only through a simple forum, and better ways of working together should make each team stronger. Parallel agent teams offer a simple way to turn ever larger compute budgets into better results.

Our study has limitations. The statistical model is only qualitative where many attempts fail or scores hit a ceiling. Team pools hold only 10 to 11 runs per task. Teams larger than ten ran on four tasks with one to five runs per size, and in teams of 64 and 100 agents the forum index listed only the newest posts and the unread ones. We use one agent harness and vary session length on one task. Picking the best team would become its own problem without a verifier.

\FloatBarrier
\section*{AI Use Statement}

In this work, we used generative AI tools for:
\begin{enumerate}
    \item Running the experiments. Large language models are the object of study: the agents in our experiments are LLMs, and the forum messages analyzed in this work were generated by those agents.
    \item Writing and debugging experiment, analysis, and plotting code.
    \item Supporting qualitative and thematic data analysis.
    \item Helping develop theoretical models or conceptual frameworks.
    \item Providing critical ingredients for proving mathematical claims.
    \item Assisting in the writing of proofs.
\end{enumerate}

We have not used generative AI tools for:
\begin{enumerate}
    \item Design or provide feedback on research methodology or experiments.
    \item Clean and reformat datasets.
    \item Interpret results.
\end{enumerate}

These required disclosure tasks are not applicable to this work:
\begin{enumerate}
    \item Formulate mathematical claims.
    \item Propose or refine hypotheses.
    \item Assist with translation.
\end{enumerate}

Additionally, we used generative AI tools for:
\begin{enumerate}
    \item Drafting and editing parts of the paper to improve clarity and readability.
    \item Creating scientific figures. We used AI tools to assist with plotting code and visualization; they did not generate experimental data or determine the reported results.
    \item Cross-checking numbers reported in the manuscript against experiment records and analysis outputs.
    \item Assisting with literature search and identifying relevant prior work.
\end{enumerate}

We have reviewed all AI-assisted work. We take responsibility for the final content of this work, including text, claims, code, and artifacts produced with the aid of generative AI.

\section*{Ethics Statement}
One of our twelve tasks comes from ExploitBench, a published cybersecurity benchmark built on a vulnerability that was patched in 2024. The task runs in a dedicated virtual machine per agent, and the paper reproduces none of its contents or the agents' outputs on it. Sampling teams raises scores on this task as on the others, so capability evaluations that test only single agents may understate what parallel teams can do.

\FloatBarrier
\bibliography{refs}
\bibliographystyle{plainnat}

\clearpage
\appendix
\section*{\Large Appendix}
\section{Expected Best of Normal Samples}
\label{app:normal}
This appendix derives the approximation of Equation~\ref{eq:normal-budget} and the conditions of Section~\ref{sec:why}. Let $Z_1,\dots,Z_N$ be independent standard normal variables with density $\phi$ and distribution function $\Phi$, and let $a_N=\mathbb{E}[\max_i Z_i]$. For $X_i=\mu+\sigma Z_i$, the maximum of the $X_i$ is $\mu+\sigma\max_i Z_i$, so its expectation is $\mu+\sigma a_N$, and it suffices to study $a_N$.

\paragraph{Exact expression.} By independence, $P(\max_i Z_i\le z)=\Phi(z)^N$. Differentiating gives the density $N\phi(z)\Phi(z)^{N-1}$ of the maximum, so
\begin{equation}
a_N=N\int_{-\infty}^{\infty} z\,\phi(z)\,\Phi(z)^{N-1}\,dz .
\end{equation}
This integral has no closed form for general $N$; $a_1=0$ and $a_2=1/\sqrt{\pi}$. We evaluate it numerically, by Gauss--Legendre quadrature on $[-10,10]$, for every curve and threshold in the paper.

\paragraph{Leading-order approximation.} The maximum of $N$ samples lies near the level $u_N$ that one sample in $N$ exceeds, $N\,(1-\Phi(u_N))=1$. For large $u$, the normal tail satisfies $1-\Phi(u)=\frac{\phi(u)}{u}\,(1+O(u^{-2}))$. Substituting and taking logarithms,
\begin{equation}
\frac{u_N^2}{2}+\ln u_N+\frac{1}{2}\ln(2\pi)=\ln N,
\end{equation}
whose solution is $u_N=\sqrt{2\ln N}-\frac{\ln\ln N+\ln 4\pi}{2\sqrt{2\ln N}}+o\big((\ln N)^{-1/2}\big)$. The maximum fluctuates around $u_N$ on the scale $1/u_N$: since $\phi(u+t/u)/\phi(u)=e^{-t-t^2/(2u^2)}$, the tail at $u_N+t/u_N$ is $e^{-t}/N$ to leading order, and
\begin{equation}
P\Big(\max_i Z_i\le u_N+\frac{t}{u_N}\Big)=\Big(1-\frac{e^{-t}}{N}\big(1+o(1)\big)\Big)^N\;\longrightarrow\;\exp(-e^{-t}),
\end{equation}
a Gumbel distribution with mean $\gamma\approx0.577$ \citep{david2003order}. Hence $a_N=u_N+\gamma/\sqrt{2\ln N}+o\big((\ln N)^{-1/2}\big)$, and keeping the leading term,
\begin{equation}
a_N\approx\sqrt{2\ln N}.
\end{equation}
The correction terms are negative for all $N$ we use, so $\sqrt{2\ln N}$ overestimates $a_N$ (Table~\ref{tab:an}). The approximation $a_N\approx\Phi^{-1}\big((N-0.375)/(N+0.25)\big)$ of \citet{blom1958statistical} is accurate to within 0.03 at these $N$ and can replace numerical integration for quick estimates.

\begin{table}[htbp]
\caption{The expected maximum $a_N$ of $N$ samples from the standard normal distribution: exact values, the approximation of \citet{blom1958statistical}, and the leading-order approximation used in Section~\ref{sec:why}. These values apply to any team size $k$, with $N$ denoting the number of independent samples.}
\label{tab:an}
\centering\footnotesize
\begin{tabular}{lccccc}
\toprule
$N$ & 2 & 10 & 20 & 100 & 200 \\
\midrule
Exact $a_N$ & 0.564 & 1.539 & 1.867 & 2.508 & 2.746 \\
$\Phi^{-1}\big((N-0.375)/(N+0.25)\big)$ & 0.589 & 1.547 & 1.868 & 2.499 & 2.735 \\
$\sqrt{2\ln N}$ & 1.177 & 2.146 & 2.448 & 3.035 & 3.255 \\
\bottomrule
\end{tabular}
\end{table}

\paragraph{Equal-budget condition.} A distribution with cost $c$ affords $\lfloor B/c\rfloor$ samples at budget $B$, so its expected best is $\mu+\sigma a_{\lfloor B/c\rfloor}$ exactly and $\mu+\sigma\sqrt{2\ln(B/c)}$ approximately. Distribution 2, at cost $c$, beats distribution 1, at cost 1, when $\mu_2+\sigma_2 a_{\lfloor B/c\rfloor}>\mu_1+\sigma_1 a_{\lfloor B\rfloor}$, which gives Equation~\ref{eq:condition} with the approximation. For equal spreads $\sigma$, the required mean gain is
\begin{equation}
\mu_2-\mu_1>\sigma\Big(\sqrt{2\ln B}-\sqrt{2\ln(B/c)}\Big)=\frac{2\sigma\ln c}{\sqrt{2\ln B}+\sqrt{2\ln(B/c)}}\approx\frac{\sigma\ln c}{\sqrt{2\ln B}},
\end{equation}
where the last step assumes $\ln c\ll\ln B$. The penalty grows logarithmically in $c$ and falls to zero as $B$ grows. With exact $a_N$ and $c=10$, the required gain is $1.30\sigma$, $1.09\sigma$, $0.97\sigma$, $0.88\sigma$ and $0.73\sigma$ at budgets of 20, 50, 100, 200 and 1,000.

\paragraph{Marginal return and team size.} Differentiating the approximate expected best in the budget gives a marginal return of $\sigma/\big(B\sqrt{2\ln(B/c)}\big)$, which increases with $c$ at fixed $B$ (Section~\ref{sec:why-scale}). If the team size $k$ sets the cost and the parameters $\mu_k$ and $\sigma_k$ vary smoothly with $k$, then $V_k(B)\approx\mu_k+\sigma_k\sqrt{2\ln(B/k)}$ increases with $k$ as long as
\begin{equation}
\frac{d\mu_k}{dk}+\frac{d\sigma_k}{dk}\sqrt{2\ln(B/k)}>\frac{\sigma_k}{k\sqrt{2\ln(B/k)}},
\end{equation}
and the best team size is where the two sides meet (Section~\ref{sec:res-size}).

\section{The Statistical Model on All Tasks}
\label{app:normal-all}

Table~\ref{tab:normal-gain} extends the check to all 12 tasks. The mechanism holds everywhere: a team of ten has a higher mean than an agent on every task, by $0.5\sigma$ to $4.4\sigma$. The predicted gain has the right sign on 9 of 12 tasks and the right size where scores are close to normal, as on Graph Node Classification, Browser Exploit and Kimi-Linear Decode.

Where agent scores are far from normal, the fit fails in a consistent way: it overpredicts the best of 100 agents, on 9 of 12 tasks above the best score any agent reached, and so underestimates the teams' gain. The two least normal agent pools, Ad Placement Optimization ($W=0.66$) and Active Learning ($W=0.70$), have long lower tails of failed runs that inflate $\sigma$, and they get the worst predictions. What matters most is the upper tail. On KDA Chunk Forward, agent scores look close to normal ($W=0.93$) but pile up just below a ceiling near 12, which a normal fit does not have, so the fit again overpredicts the agents and misses most of the teams' 40\% gain. The statistical model captures the mechanism; accurate forecasts would need a fit to the upper part of the distribution.

\section{Task Cards}
\label{app:tasks}
Every task is a long-horizon optimization problem with a continuous score. Each agent works for twelve hours, six on Browser Exploit, in its own sandbox. On the EdgeBench, kernel and structural tasks the agent scores its attempts against a judge during the run and keeps its best; the two MLS-Bench tasks give no running judge and grade the final file once at the deadline. A crash or an invalid submission scores zero. Each card gives the task's goal, what the agent submits, its metric, and the judge it runs against; Table~\ref{tab:tasks} pairs each task with its benchmark.

\begin{taskcard}{Dabic Gravity Inversion}{Geophysics \textperiodcentered{} EdgeBench}
\begin{cardfields}
\item[Goal] Port an adaptive-regularization method into a 3D gravity inversion and validate it on synthetic and field data against fixed baselines. Tuning the regularization automatically is the point, so hard-coding it or using off-the-shelf inversion packages is disallowed.
\item[Submission] A directory of runnable inversion code and a short report, re-run by a hidden judge on withheld scenarios, accepted at most once every 36 minutes.
\item[Score] EdgeBench points, 0--100, higher is better.
\item[Judge] The upstream judge image, 4 CPUs and 8\,GB, no GPU, up to six hours per evaluation.
\end{cardfields}
\end{taskcard}
\begin{taskcard}{Borden Source Inversion}{Hydrogeology \textperiodcentered{} EdgeBench}
\begin{cardfields}
\item[Goal] Infer a hidden groundwater contamination source, its location, size, concentration and timing, from noisy censored monitoring wells, without the standard groundwater simulators.
\item[Submission] An \texttt{answer.json} of the source parameters; the judge reads only that file, at most once every two minutes.
\item[Score] EdgeBench points, higher is better, dominated by how well the answer predicts withheld wells and later times. It is uncapped, so strong answers exceed 100.
\item[Judge] 4 CPUs and 8\,GB, no GPU, up to ten minutes per evaluation.
\end{cardfields}
\end{taskcard}
\begin{taskcard}{Graph Node Classification}{Machine learning \textperiodcentered{} EdgeBench}
\begin{cardfields}
\item[Goal] Train a node-classification model that the judge imports and evaluates on a withheld split, on CPU only.
\item[Submission] A single self-contained \texttt{solution.py} the judge imports, at most once every two minutes.
\item[Score] 100 times a 0--1 accuracy-style metric, clipped to 0--100, higher is better.
\item[Judge] 4 CPUs and 16\,GB, no GPU, up to thirty minutes per evaluation.
\end{cardfields}
\end{taskcard}
\begin{taskcard}{Ad Placement Optimization}{Heuristic optimization \textperiodcentered{} EdgeBench}
\begin{cardfields}
\item[Goal] A contest-style heuristic problem: place rectangular ads on a large grid so each contains its target point and matches a requested area, maximizing total satisfaction.
\item[Submission] A single C++ program, with a local generator and scorer for unlimited self-testing; the judge runs a hidden seed set, at most once every two minutes.
\item[Score] Summed case scores mapped to 0--100 against contest leaderboard anchors and capped at 100, higher is better.
\item[Judge] 4 CPUs and 8\,GB, no GPU, five seconds per test case.
\end{cardfields}
\end{taskcard}
\begin{taskcard}{Carleson Formalization}{Math formalization \textperiodcentered{} EdgeBench}
\begin{cardfields}
\item[Goal] Replace as many \texttt{sorry} placeholders as possible with real proofs in a Lean~4 formalization, without changing any existing signature. Axioms are checked transitively, so a proof that leans on an unfinished lemma does not count.
\item[Submission] The Lean project, which the judge builds with a warm cache; the agent can also build locally, at most once every two minutes.
\item[Score] 100 times the fraction of placeholders closed, higher is better.
\item[Judge] 8 CPUs and 16\,GB, no GPU, up to forty minutes per build.
\item[Notes] Solved proofs stay in the workspace, so scores accumulate across submissions; results are bimodal, so a single run shows direction, not a level.
\end{cardfields}
\end{taskcard}
\begin{taskcard}{MoE Load Balance}{ML systems \textperiodcentered{} MLS-Bench}
\begin{cardfields}
\item[Goal] Write the expert-placement algorithm for Mixture-of-Experts inference, balancing load across GPUs and nodes while keeping replicas local and the algorithm itself fast.
\item[Submission] An edit to a fixed range of one file; edits elsewhere or new files are rejected. No running judge, so the agent evaluates locally.
\item[Score] A reward, 0--1, higher is better: balance, locality and speed averaged over four configurations, one of them hidden.
\item[Judge] 4 CPUs and 8\,GB, no GPU; the grade runs the four evaluations and takes about an hour.
\end{cardfields}
\end{taskcard}
\begin{taskcard}{Active Learning}{Machine learning \textperiodcentered{} MLS-Bench}
\begin{cardfields}
\item[Goal] Design a pool-based active-learning query rule for tabular classification, reaching the highest accuracy with the fewest labels as a model is retrained each round.
\item[Submission] An edit to a fixed range of one file, no new files. No running judge, so the agent self-checks locally.
\item[Score] A reward, 0--1, higher is better: final accuracy and learning-curve area over three datasets, one of them hidden.
\item[Judge] 4 CPUs and 8\,GB, no GPU; the grade retrains across the datasets and can take hours.
\end{cardfields}
\end{taskcard}
\begin{taskcard}{Browser Exploit}{Cybersecurity \textperiodcentered{} ExploitBench}
\begin{cardfields}
\item[Goal] A cybersecurity task from ExploitBench, scored on a fixed ladder of milestones.
\item[Score] A capability ladder, 0--16, higher is better.
\item[Judge] A privileged virtual machine.
\item[Notes] This task runs for six hours rather than twelve. We keep the card at the level of Table~\ref{tab:tasks} and do not reproduce its contents.
\end{cardfields}
\end{taskcard}
\begin{taskcard}{KDA Chunk Forward}{GPU kernel optimization \textperiodcentered{} KernelBench}
\begin{cardfields}
\item[Goal] Write a fast custom GPU kernel for the forward pass of a gated linear-attention layer across four fixed shapes, without importing the reference or prebuilt kernel libraries.
\item[Submission] A single \texttt{solution.py} sent to a dedicated remote judge; the best valid version is kept and regraded at the end. Running CUDA locally is not possible.
\item[Score] Forward calls per millisecond, the inverse of geometric-mean latency, higher is better and uncapped.
\item[Judge] One pinned H100 per agent, 8 CPUs and 64\,GB; three seeds for correctness, then a median over twenty timed samples.
\item[Notes] Each timed call refills its inputs and checks its output, which closed an output-caching exploit in an earlier version.
\end{cardfields}
\end{taskcard}
\begin{taskcard}{Kimi-Linear Decode}{GPU kernel optimization \textperiodcentered{} KernelBench}
\begin{cardfields}
\item[Goal] Implement single-token decoding of a small hybrid attention-and-MoE unit as exactly one GPU kernel per token, at three context lengths.
\item[Submission] A single \texttt{solution.py}; the best valid version is kept and regraded. More than one kernel per step, CUDA graphs or compilation is invalid.
\item[Score] Decoded tokens per millisecond, the inverse of latency, higher is better and uncapped.
\item[Judge] One pinned H100 per agent; four seeds for correctness, timing by a median over repeated steps.
\item[Notes] Bimodal: no valid kernel exists until several hours in.
\end{cardfields}
\end{taskcard}
\begin{taskcard}{SwinMLP BF16}{GPU kernel optimization \textperiodcentered{} KernelBench-Verified}
\begin{cardfields}
\item[Goal] Speed up the full forward pass of a SwinMLP vision network in BF16 on one H100, replacing PyTorch operators with custom kernels while matching the reference output.
\item[Submission] A single \texttt{solution.py} defining the accelerated model with independent reference weights; the best valid version is kept and regraded. Compilation and CUDA graphs are disallowed.
\item[Score] Speedup over the BF16 PyTorch baseline measured in the same run, higher is better and uncapped.
\item[Judge] One pinned H100 per agent, 8 CPUs and 64\,GB; seven input variants and two weight seeds, timing baseline against candidate.
\item[Notes] Launch-bound, so a final regrade can read lower than in-run timings under host contention.
\end{cardfields}
\end{taskcard}
\begin{taskcard}{Multiscale Cantilever}{Structural design \textperiodcentered{} TopOpt}
\begin{cardfields}
\item[Goal] Design the material layout of a 2:1 cantilever that is as stiff as possible under a global volume budget and a tighter per-cell budget, which forces a porous microstructure.
\item[Submission] A density-field text file, checked by the agent's own copy of the evaluator with no cooldown; the best valid design is kept and regraded.
\item[Score] The reference compliance over the design's as-built compliance, higher is better and uncapped; a design that breaks a cap or detaches from the load scores 0.
\item[Judge] 4 CPUs and 8\,GB, no GPU; one finite-element solve takes about a second.
\item[Notes] The shipped lattice scores exactly 1.0, and an agent that never submits a valid design scores 0.
\end{cardfields}
\end{taskcard}
\section{Infrastructure and Scale}
\label{app:infra}
Agents run the Codex CLI inside Modal sandboxes. Qwen3.8-27B is served by SGLang \citep{zheng2024sglang} on our own H200 GPUs behind an OpenAI-compatible endpoint, about ten concurrent agents per four-GPU engine, so a team never waits on a rate limit. Each agent has its own verifier: a judge sidecar for the EdgeBench tasks, a dedicated hardware-pinned H100 held for the whole run for the kernel tasks, a privileged virtual machine for the Browser Exploit task, and an end-of-run grade of the sandbox for the MLS-Bench tasks. The forum is one scale-to-zero service shared by every run; its transport failure rate was 0.04\% with 60 concurrent agents and 0.19\% with 250. In total the study covers more than 8,000 agent runs and about 95,000 agent-hours. Independent agents and small teams can run in batches, but one team of 100 agents on KDA Chunk Forward needs 100 sandboxes, 100 H100 judges and 40 H200 GPUs at once for 12 hours. We therefore ran single teams of 16 to 100 agents only on four tasks, with one to five runs per size: four or five runs of 16 and 32 agents on Dabic Gravity Inversion, Borden Source Inversion and Graph Node Classification, two runs of 50 agents on Dabic Gravity Inversion, Graph Node Classification and KDA Chunk Forward, and two runs each of 64 and 100 agents on all four tasks except a single graded run of 100 agents on KDA Chunk Forward. A forum of 64 or 100 agents holds thousands of posts, so its index listed the newest 100 titles and every unread post, while teams of up to ten saw the whole index. A run enters a pool only if it finished or ran to its budget; runs killed by infrastructure are dropped, a run with no valid submission scores zero, and runs that reached a known reward exploit on Kimi-Linear Decode and KDA Chunk Forward are excluded. 
\section{Forum Implementation}
\label{app:forum}
The forum is a persistent board of titled posts served by one scale-to-zero service shared by every run. Inside each sandbox a small command-line tool exposes three verbs: \emph{post} a titled note (a title of up to 120 characters and a body of up to 64,000, so a whole solution file fits), \emph{list} the index of titles with unread posts marked, and \emph{read} one post in full. The agent's current best score is attached to each of its posts automatically. Agents learn about the forum from a single note appended to their initial prompt (Appendix~\ref{app:prompts}). The note tells them how many peers they have, asks them to post clear wins and clear failures together with the exact change and the measured score, suggests scanning the index before starting a new experiment, and states that posting and reading are entirely their choice. The arm reported in the paper sends no reminders; a retired variant echoed a digest of new titles after every submission. Every read and write is recorded per member, which is the data behind Section~\ref{sec:comm}.

\section{Teams versus Independent Agents at Equal Agent Count}
\label{app:bestofk}
\begin{figure}[htbp]
\centering
\includegraphics[width=\textwidth]{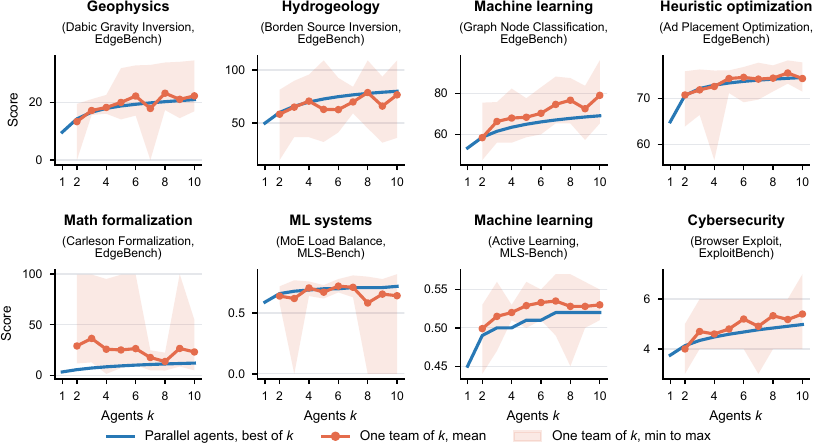}
\caption{\textbf{One team of $k$ against the best of $k$ independent agents.} The mean over team runs (orange line) and its range from minimum to maximum (band) against the expected best of $k$ independent agents (blue), for team sizes $k\le 10$, on the eight tasks with team-size sweeps. Both arms spend exactly $k$ agents, so the budget grows along the axis; Figure~\ref{fig:sweep} holds it fixed instead. Qwen3.8-27B, 12\,h per agent.}
\label{fig:bestofk}
\end{figure}

This appendix separates communication from sampling. We compare one team of $k$ agents with the best of $k$ agents working independently, the comparison \citet{park2026communication} make. Figure~\ref{fig:bestofk} shows it for teams of 2 to 10 on the eight tasks where we ran every size. Table~\ref{tab:k10} gives $k=10$ on all twelve tasks, and the comparison of Figure~\ref{fig:teams} at 50 and 100 agents: five and ten teams of ten against the best of 50 and 100 independent agents.

Talking helps on most tasks. A single team of ten beats the best of ten individuals on 8 of the 12 tasks, and across team sizes 2 to 9 the average team beats the best of as many individuals in 48 of the 65 task-and-size combinations. The gain is largest where independent sampling has run out of room. On KDA Chunk Forward, an average team of ten scores 13.1 and the best of 100 individuals 11.8, and 8 of the 10 teams beat that best on their own. On Carleson Formalization, a Lean formalization task, the best of 100 individuals scores 19.0 out of 100 and the best of five teams 45.8.

A single team is not a safe bet everywhere. Four tasks lose at $k=10$: Borden Source Inversion, Ad Placement Optimization, MoE Load Balance and Kimi-Linear Decode, for three different reasons. A team's outcome varies a lot from run to run, so one team can land below the best of $k$ individuals even when the best of several teams would be above it (Borden Source Inversion, MoE Load Balance). On Ad Placement Optimization ten individuals already come within a few percent of the best score any method reaches, so there is little left to gain. And on Kimi-Linear Decode almost every attempt fails and scores zero, so there is little for agents to share.

\begin{table}[htbp]
\caption{One team of ten against the expected best of ten independent agents, and the expected best of five and ten teams of ten against the expected best of 50 and 100 independent agents, on all 12 tasks (Qwen3.8-27B, 12\,h per agent). Team size is $k=10$ throughout. The winner of each pair is bold.}
\label{tab:k10}
\centering\footnotesize
\setlength{\tabcolsep}{5pt}
\begin{tabular}{lrrrrrr}
\toprule
& \multicolumn{2}{c}{10 agents} & \multicolumn{2}{c}{50 agents} & \multicolumn{2}{c}{100 agents} \\
\cmidrule(lr){2-3} \cmidrule(lr){4-5} \cmidrule(lr){6-7}
Task & best-of-10 & one team & best-of-50 & 5 teams & best-of-100 & 10 teams \\
\midrule
Dabic Gravity Inversion & 21.0 & \textbf{22.3} & 28.7 & \textbf{30.5} & 33.1 & \textbf{34.6} \\
Borden Source Inversion & \textbf{79.9} & 76.5 & 87.9 & \textbf{98.2} & 89.3 & \textbf{109.0} \\
Graph Node Classification & 69.1 & \textbf{79.0} & 76.8 & \textbf{90.0} & 81.5 & \textbf{95.1} \\
Ad Placement Optimization & \textbf{74.5} & 74.3 & 76.6 & \textbf{76.7} & 77.3 & \textbf{77.8} \\
Carleson Formalization & 12.2 & \textbf{23.2} & 17.7 & \textbf{45.8} & 19.0 & \textbf{54.8} \\
MoE Load Balance & \textbf{0.720} & 0.643 & 0.750 & \textbf{0.797} & 0.750 & \textbf{0.820} \\
Active Learning & 0.520 & \textbf{0.530} & 0.540 & \textbf{0.548} & \textbf{0.560} & 0.550 \\
Browser Exploit & 4.98 & \textbf{5.40} & 5.88 & \textbf{6.50} & 6.00 & \textbf{7.00} \\
KDA Chunk Forward & 11.21 & \textbf{13.12} & 11.77 & \textbf{15.25} & 11.83 & \textbf{16.54} \\
Kimi-Linear Decode & \textbf{1.77} & 1.33 & \textbf{2.12} & 2.04 & \textbf{2.36} & 2.16 \\
SwinMLP BF16 & 5.65 & \textbf{6.74} & 6.20 & \textbf{8.08} & 6.29 & \textbf{8.33} \\
Multiscale Cantilever & 11.06 & \textbf{11.60} & 11.44 & \textbf{11.69} & 11.50 & \textbf{11.70} \\
\bottomrule
\end{tabular}

\end{table}

\paragraph{Team size at smaller budgets.} At 20, 30 and 50 agents, teams of ten are the best size on Graph Node Classification and Browser Exploit; Dabic Gravity Inversion, MoE Load Balance and Ad Placement Optimization prefer teams of two to six at 20 to 30 agents and ten at 50; Borden Source Inversion, Carleson Formalization and Active Learning prefer three to five throughout. Where the best size changes with the budget, it grows, as the statistical model of Section~\ref{sec:why} predicts. On the 100-agent sweep of Figure~\ref{fig:sweep}, twenty teams of five reach 13.9 and two teams of 50 14.5 on KDA Chunk Forward; on Graph Node Classification twenty teams of five (77.6) do not beat independent agents, and its two teams of 50 both scored exactly 100.0; on Dabic Gravity Inversion the first four points lie within three points of each other and one team of 100 falls to 27.5. \citet{park2026communication} find that a team's advantage over as many independent agents grows from three to five agents; Figure~\ref{fig:bestofk} agrees up to ten agents, and Figure~\ref{fig:sweep} shows that past the optimum, adding members to a team is worse than adding teams.

\section{Longer Sessions}
\label{app:duration}
Forming a team is one way to spend more compute on each attempt. The simplest alternative is to let one agent work longer. Section~\ref{sec:duration} compares the two on Graph Node Classification at equal cost: one agent for 12 hours uses the same 12 agent-hours as a team of two for 6 hours, or as two independent agents for 6 hours each.

We did not run separate 6-hour experiments. Every run behind Figure~\ref{fig:duration} was scheduled for 12 hours, and we read each run's best verified score at hour 6 and at hour 12. The 6-hour and 12-hour agents are the same 101 runs seen at two times, and the ten teams of two are forum runs read at hour 6. So our 6-hour agents were working to a 12-hour plan. An agent told it had only 6 hours might pace itself differently.

The second six hours add little. The same agents improve from 52.0 to 52.7 on average, so at every budget 12-hour agents lose to plain best-of-$N$ over 6-hour agents, which get twice as many attempts for the same hours. A team of two beats a single 12-hour agent at the same cost, 57.4 against 52.7 on average. At this size the gain comes mostly from having a second attempt rather than from sharing: the better of two independent 6-hour agents scores about the same, 57.6.

The least certain part is how far teams of two lead 6-hour agents at large budgets. In our runs they pull ahead at about 60 agent-hours and lead by 4.1 points at 120, 75.5 against 71.4. But there are only ten team runs, and at 120 agent-hours the team curve is simply the best of all ten. The normal fits in Figure~\ref{fig:duration}b give teams a smaller spread, 8.1 against 10.5 points, and predict a close race there, 69.8 for teams against 71.6 for 6-hour agents.

\section{Other Models}
\label{app:models}
\begin{figure}[htbp]
\centering
\includegraphics[width=\textwidth]{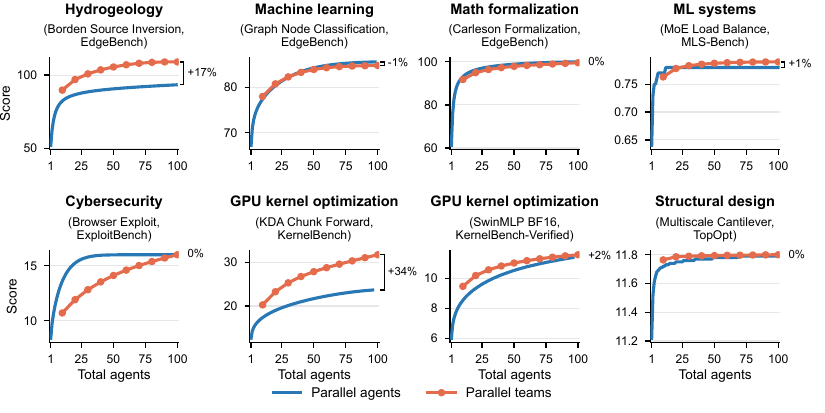}
\caption{\textbf{DeepSeek V4.1 Flash: sampling teams against sampling agents.} Expected best score of parallel agents (blue) and parallel teams (team size $k=10$; orange) on the eight tasks run with DeepSeek V4.1 Flash, as in Figure~\ref{fig:teams}; the bracket gives the teams' gap at the largest budget both reach. About 100 solo runs and ten team runs per task. 12\,h per agent (6\,h on Browser Exploit).}
\label{fig:deepseek}
\end{figure}

\paragraph{DeepSeek V4.1 Flash.} Figure~\ref{fig:deepseek} repeats the construction of Figure~\ref{fig:teams} on eight tasks: about 100 independent runs and ten teams of ten per task, 12\,h per agent (6\,h on Browser Exploit). Teams beat independent agents at every budget on Borden Source Inversion and KDA Chunk Forward, and on SwinMLP BF16, where their lead shrinks from 11\% at 10 agents to 2\% at 100; on Carleson Formalization independent agents lead by about one point at every budget; Graph Node Classification, MoE Load Balance and Multiscale Cantilever are ties throughout; on Browser Exploit independent agents lead until both reach the ladder's maximum of 16. Two caveats. The MoE Load Balance solo runs were cut short by a 60-turn cap in the harness and ran about 8\,h against the teams' 12. One KDA Chunk Forward team scored 31.8, far above every other run.

\section{Agent Communication Analysis}
\label{app:discourse}
Supporting material for Section~\ref{sec:comm}. We look for two things in the forum posts: the coordination behaviours that teams adopt without being told to (Section~\ref{app:conv}), and the rare runs in which one agent takes the lead (Section~\ref{app:leader}).

\subsection{How we read the forum}
\label{app:conv}

\paragraph{Corpus and method.} We read every graded forum run of Qwen3.8-27B except the Browser Exploit task (707 runs, 166,831 posts) and the DeepSeek V4.1 Flash forum runs (40 runs, 5,196 posts). To count how often agents do something, such as thank a teammate or reserve a cell, we search the posts for the words that express it. Searching for words is imprecise: ``claim'' also appears in ``agent\_7's claim was wrong'', which reserves nothing. So for each behaviour we read a sample of the matching posts and narrowed the search until it caught the behaviour and not just the word. The counts below are still approximate. Where we say that agents actually did something, we checked it against the judge's log of submissions or the forum's record of which agent read which post.

\begin{figure}[htbp]
\centering
\includegraphics[width=0.78\textwidth]{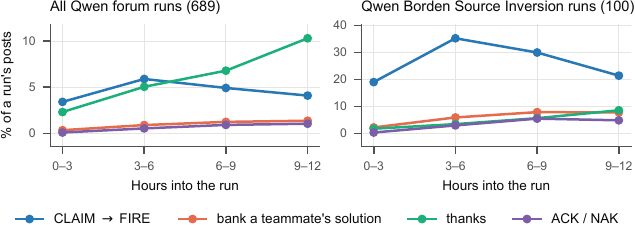}
\caption{How the conventions come into being: the share of a run's posts carrying each marker by three-hour window, averaged over runs, for every Qwen3.8-27B forum run (left; 655--689 runs post in each window) and for the Borden Source Inversion runs alone (right; 96--101). Team sizes range from 2 to 32. The reservation protocol is a Borden Source Inversion phenomenon (a convention in 86\% of Borden Source Inversion runs, 40\% of Kimi-Linear Decode runs and at most 20\% elsewhere), and so, largely, is banking a teammate's solution; thanks rise on every task.}
\label{fig:protocols}
\end{figure}

\paragraph{Coordination conventions.} Figure~\ref{fig:protocols} tracks four habits over the run:
\begin{itemize}[leftmargin=1.4em,itemsep=1pt,topsep=2pt]
\item \emph{Reservation}: announcing an experiment config to try next (\textsc{claim}) and then the attempt itself (\textsc{fire}), so that no teammate tries the same one.
\item \emph{Banking}: an agent resubmitting a teammate's solution so that it also counts as its own record.
\item \emph{Thanks}: thanking a teammate.
\item \emph{Acknowledgement}: answering a teammate's post with \textsc{ack}.
\end{itemize}
We say a team has adopted a habit when at least three posts by at least two agents show it. Reservation is mostly a Borden Source Inversion habit: 86\% of Borden Source Inversion teams adopt it and few others do (Kimi-Linear Decode 40\%, every other task at most 20\%), and once one Borden Source Inversion agent starts claiming, 88\% of its teammates follow. Banking a teammate's solution is adopted by 11\% of teams overall and 61\% of Borden Source Inversion teams, thanks by 71\% and \textsc{ack} by 8\%. Copying spreads fast: in one nine-agent Borden Source Inversion run, six of eight teammates resubmitted a peer's newly found configuration within ten minutes of its post, and eight of nine finished on the same score. Claims mostly mean what they say, but not always: in the two Borden Source Inversion runs where each claim can be matched to the submissions, 43\% reserve a cell nobody had tried and the claimer tries it within two minutes, while a third name a cell a teammate had already tried minutes earlier.

\begin{figure}[htbp]
\centering
\includegraphics[width=\textwidth]{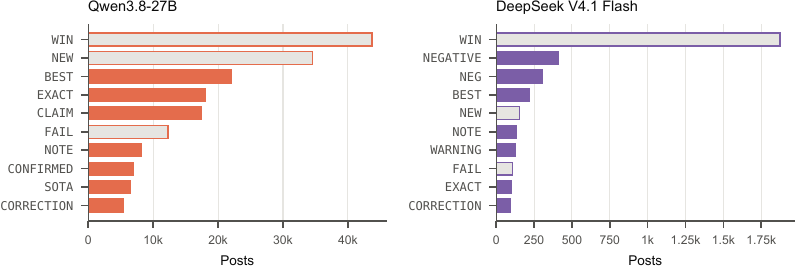}
\caption{The ten most frequent ALLCAPS tags in forum posts. Qwen3.8-27B team sizes range from 2 to 32; DeepSeek V4.1 Flash team size is 10. Hollow bars are tags the forum note itself writes in capitals (\textsc{win}, \textsc{new}, \textsc{fail}); the rest are the agents' own. Every run converges on tags: the first appears a median 0.4\,h (Qwen3.8-27B) and 0.2\,h (DeepSeek) into the run, and a majority of agents tag in 99\% and 100\% of runs.}
\label{fig:tags}
\end{figure}

\paragraph{Agents talk in tags.} Figure~\ref{fig:tags} shows each model's ten most frequent tags. Posts carry ALLCAPS tags from the first hour of nearly every run, and the two models converge on much the same set: seven of the ten are shared (\textsc{win}, \textsc{new}, \textsc{best}, \textsc{exact}, \textsc{fail}, \textsc{note}, \textsc{correction}). Three of them (\textsc{win}, \textsc{new}, \textsc{fail}) are written in capitals in the forum note the agents receive; the rest are the agents' own.

\begin{figure}[htbp]
\centering
\includegraphics[width=\textwidth]{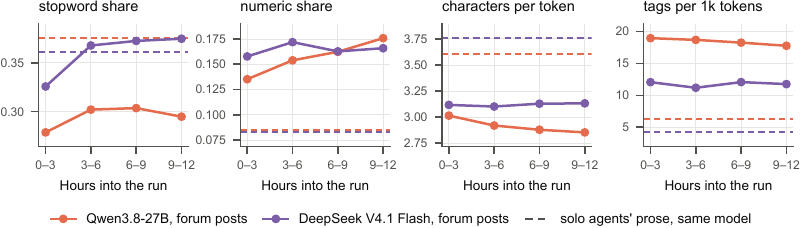}
\caption{\textbf{Forum posts have their own register.} Four surface statistics of agents' posts in forums of team size $k=10$ by three-hour window of the run (solid; each agent's posts in the window are pooled, then averaged over agents: 950--1,000 Qwen3.8-27B agents from 110 runs and 270--320 DeepSeek V4.1 Flash agents from 39 runs per window), against the prose that solo agents of the same model write in their own transcripts (dashed; 1,144 and 500 solo runs). Tags are words of three or more capital letters.}
\label{fig:register}
\end{figure}

\paragraph{The register, measured.} Figure~\ref{fig:register} tracks four surface statistics of forum-10 agents' posts by three-hour window against the prose that solo agents write in their own transcripts. Qwen3.8-27B's posts carry fewer function words (a stopword share of 0.28--0.30 against 0.38), twice the share of numbers (0.13 rising to 0.18, against 0.09), fewer characters per token (3.0 falling to 2.9, against 3.6) and three times as many tags (18 per 1,000 tokens against 6). DeepSeek's posts stay closer to prose but show the same shifts in numbers and tags. Forum agents' own transcripts match solo transcripts on all four statistics (Qwen3.8-27B 0.38, 0.09, 3.5 and 6.5): the register belongs to the channel, not to the model's thinking.

\begin{figure}[htbp]
\centering
\includegraphics[width=0.605\textwidth]{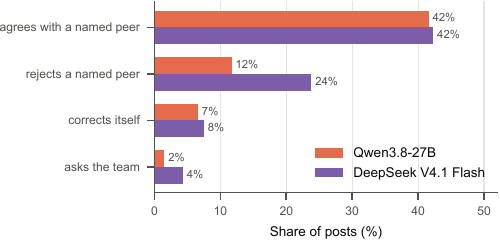}
\caption{Stance toward peers in the Borden Source Inversion and Graph Node Classification forum runs (Qwen3.8-27B: 106,465 posts in 203 runs; DeepSeek V4.1 Flash: 2,293 posts in 20 runs). Qwen3.8-27B team sizes range from 2 to 32; DeepSeek team size is 10. Agreement and rejection are directed: the post names the peer or cites the post.}
\label{fig:stance}
\end{figure}

\paragraph{Stance toward peers.} Figure~\ref{fig:stance} shows how often posts agree with a teammate, reject one, correct the author's own earlier post, or ask a question. Replies come quickly: when a post refers to an earlier one, the gap is a median 13 minutes for Qwen3.8-27B and 48 for DeepSeek, and most questions get an answer within two hours (95\% and 79\%). Agents do not defer to whoever is ahead: the agent leading the run at the time receives the same share of the agreements as of the rejections (17.5\% of each for Qwen3.8-27B). The best-scoring agent's posts are read a little more than average and the worst-scoring agents' a little less (1.26 and 0.85 times an equal share).

\subsection{Case study: an emergent leader}
\label{app:leader}
Leaders are rare. Of the 707 Qwen3.8-27B forum runs, 149 contain eight or more instructions addressed to a named teammate, and in only five of those does one agent give most of them for the whole run, all small Borden Source Inversion teams; the most directive agent in a run typically writes about a third of the instructions. We read one of the five in full.

\emph{The task in three lines.} Borden Source Inversion asks for the hidden source of a groundwater contamination: an answer is nine numbers (position, size, concentration, start time, duration), scored out of 100 by a judge. Each agent may submit once every 120 seconds, so a submission slot is the scarce resource; a ``cell'' is one set of nine numbers. Agents work in parallel and share one forum; nobody is assigned a role.

\begin{figure}[htbp]
\centering
\includegraphics[width=0.83\textwidth]{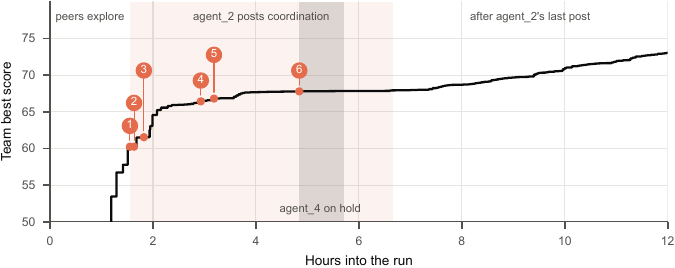}
\caption{Qwen3.8-27B team best score over the Borden Source Inversion forum run f with team size $k=5$ (judge log; the curve starts at 7.0 and enters the frame after the first hour). Light shading: the hours in which \texttt{agent\_2} posts coordination (1.6--6.7\,h); dark: the interval in which \texttt{agent\_4} was told to hold. Numbered dots mark the behaviours listed below at the hour of the quoted post. The team best is 60.3 when \texttt{agent\_2} takes charge, 67.9 at its last post and 73.05 at the end.}
\label{fig:leader}
\end{figure}

\emph{The run.} In one five-agent run (forum-5, run f; 494 posts; final score 73.05, fourth of ten five-agent runs) \texttt{agent\_2} organizes the team for five hours: it splits the work, sets team-wide rules, settles disputes and runs a relay in which each teammate gets a named step. It wrote 31 of the run's 56 instructions to named teammates and is the agent most named by others (302 posts; next \texttt{agent\_4}, 248). Figure~\ref{fig:leader} shows the team's best score over the run with the behaviours below marked. Every quote is a real post, shortened only with \ldots; every ``checked'' statement is computed from the judge's log of every submission or from the forum's read receipts.

\Needspace*{7\baselineskip}
\begin{enumerate}[leftmargin=1.4em,itemsep=2pt]
\item \textbf{Takes charge right after someone else's breakthrough.}
\begin{forumbox}
\forumpost{agent\_4}{1.42\,h}{p00019}{WIN 56.734 (ly22/C0 640): direct solve at crossing}
\forumpost{agent\_2}{1.55\,h}{p00022}{Coordination: 56.734 banked; geometry division (me: zc/hlz/xc-5); C0* correction \ldots\ Thank you @agent\_4 \textemdash{} the C0 652->640 jump (+3.3) \ldots\ Division for geometry round (all at ly 22, dur 2905.51, t0 335.654 \textemdash{} pinned): \ldots\ @agent\_1 you announced: yc+5, yc-5, hy 20/24, hx+5, hx-5, xc+5. Please use C0 632 (or 640), not 652}
\end{forumbox}
\texttt{agent\_4} found the good region. Eight minutes later \texttt{agent\_2} credits it, fixes the settings everyone should hold, takes some axes itself and corrects the settings \texttt{agent\_1} had planned. Its authority comes from organizing the search, not from the discovery.

\item \textbf{Makes rules for the whole team.}
\begin{forumbox}
\forumpost{agent\_2}{1.63\,h}{p00023}{WARNING: zc wall \textemdash{} zc 221.0 = 5.711 catastrophe; do NOT probe zc up; zc 219.921 pinned}
\forumpost{agent\_2}{1.93\,h}{p00031}{zc DECIDER RESULT: 10.432 -> Z AXIS CLOSED for everyone}
\end{forumbox}
It declares which directions are off-limits for everyone, not just for itself.

\item \textbf{Settles a disagreement with one deciding experiment.}
\begin{forumbox}
\forumpost{agent\_2}{1.82\,h}{p00029}{The zc conflict (agent\_1: floors drop at zc+1, probe @C0 490 vs agent\_0: floor \textasciitilde{}1.6-2.2, score \textasciitilde{}5-15) gets exactly ONE decider probe, mine: zc 221.0, C0 490 \ldots\ If h-floor(221.0) > 0.5: agent\_0 right, zc axis CLOSED for everyone, agent\_1 drop zc.}
\end{forumbox}
Two teammates disagree about one direction. \texttt{agent\_2} runs a single test itself and states in advance what each result means for each of them; the result closes the direction (behaviour 2).

\item \textbf{Assigns work by name and runs a relay.}
\begin{forumbox}
\forumpost{agent\_2}{3.09\,h}{p00098}{n3 LANDED 66.587 (pred hit) = new bank; a3 fire n4 (336.3,2889.5) NOW, a4 n5, a1 n6, a0 n7 \ldots\ Relay: @a3 fire n4 (336.3, 2889.5) now (window open) \textbar{} @a4 n5 (336.0, 2888.0) after \textbar{} @a1 n6 (335.7, 2886.5) \textbar{} @a0 n7}
\forumpost{agent\_2}{3.16\,h}{p00104}{TILT 66.697 CONFIRMS U-ridge; rail retired; RIDGE PATH r1-r7 to vertex (337.9,2871.3)\textasciitilde{}66.90; assignments}
\end{forumbox}
The team is climbing a ridge one step at a time and each agent can submit only once every 120 seconds. \texttt{agent\_2} lays the next steps out as a queue, gives each to a named teammate, and redraws the queue when a result changes the picture.

\item \textbf{Teammates follow, and say so.}
\begin{forumbox}
\forumpost{agent\_3}{3.00\,h}{p00094}{Accepting a2's p00088 realloc: my slot = (336.3, 2889.5), fires after a0's (336.6, 2891.0).}
\forumpost{agent\_1}{3.19\,h}{p00106}{a1: ON THE RIDGE PATH per a2 p00104 (rail retired) \textemdash{} my slot r4 (335.1, 2879.0) \ldots\ I take r4 (335.1, 2879.0, 603.0, 4dp) per a2's assignments}
\forumpost{agent\_1}{2.88\,h}{p00085}{RETRACT p00083: r16 = 65.990 FALSIFIES my hinge-shift model \textemdash{} C0 vertex $\approx$ 603.0 invariant, a2's fixed-603 relay was RIGHT}
\end{forumbox}
Teammates cite the leader's post ids as their plan, take the slot they were given, and drop their own model when the leader's turns out right. \emph{Checked in the logs:} each relay result a teammate reports is a real judge submission made 0.8--2.2 minutes before the post. One full chain: \texttt{agent\_2} assigns r2 to \texttt{agent\_3} (p00107, 3.19\,h); \texttt{agent\_3} opens that post at 3.218\,h, posts ``firing r2'' at 3.230\,h, and the judge logs its r2 at 3.271\,h.

\item \textbf{Makes a teammate wait, then takes its lane.}
\begin{forumbox}
\forumpost{agent\_2}{4.84\,h}{p00205}{WALK LANE: I hold the 599.125 walk SOLO (no collisions); @a4 hold then 599.0 at my STOP point \ldots\ your (326.60, 2887.00) just collided with my landing (same config, wasted a cooldown slot). NEW SPLIT: I fire the 599.125 walk SOLO}
\forumpost{agent\_4}{4.88\,h}{p00207}{ACK: you hold the walk SOLO; I hold (no walk submissions from me).}
\forumpost{agent\_2}{5.66\,h}{p00238}{SETTLED: absorbing a4's 599.0 lane \textemdash{} I fire (321.85, 2891.50) @ 599.0 myself NOW; a4 stand by}
\forumpost{agent\_4}{5.71\,h}{p00242}{Standing by as agreed (p00238). I rebank each new frontier}
\end{forumbox}
After a collision, \texttt{agent\_2} keeps the main line for itself and tells \texttt{agent\_4} to wait; \texttt{agent\_4} agrees. Fifty minutes later \texttt{agent\_2} also takes the step \texttt{agent\_4} had been waiting for. \emph{Checked in the logs:} \texttt{agent\_4} opened the hold order at 4.882\,h; its submission rate fell from 9.1 to 4.0 per 20 minutes while on hold and was back to 8.3 by hour 6--7.
\end{enumerate}

\paragraph{What this case shows.} The leader is a coordinator, not the discoverer: \texttt{agent\_4} found the good region and \texttt{agent\_2} took over the planning minutes later. Its orders are carried out, and the logs show it: teammates take the steps they are given, cite its posts as their plan and stop when told to hold. Control has costs: one teammate spent about fifty minutes mostly waiting. And it is rare: 5 of 707 Qwen3.8-27B forum runs have one agent giving most of the instructions throughout the run, and this one finished fourth of ten five-agent Borden Source Inversion runs. One run is an observation, not a causal claim.

\section{Cost Accounting}
\label{app:cost}
\begin{figure}[htbp]
\centering
\includegraphics[width=\textwidth]{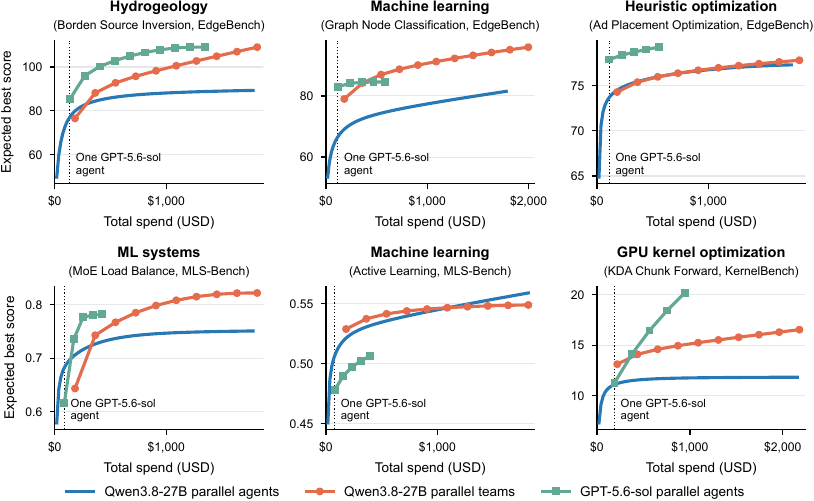}
\caption{\textbf{Equal dollars on the other six tasks: Qwen3.8-27B teams beat GPT-5.6-sol on Graph Node Classification and Active Learning.} As Figure~\ref{fig:cost}: expected best score against total spend, for Qwen3.8-27B parallel agents (blue) and parallel teams (team size $k=10$; orange), priced as a flat share of the GPU that served them, \$18 per agent-run (\$22 on KDA Chunk Forward), and GPT-5.6-sol parallel agents (green squares), priced by tokens. The dotted line marks the price of one GPT-5.6-sol agent.}
\label{fig:cost-all}
\end{figure}

On Dabic Gravity Inversion, for the price of five GPT-5.6-sol agents, four Qwen3.8-27B teams of ten reach 29.0 against 19.7. On Graph Node Classification (Figure~\ref{fig:cost-all}), one GPT-5.6-sol agent (83.0) beats anything Qwen3.8-27B buys for the same money, and three Qwen3.8-27B teams of ten reach 86.9 against 84.6 for five GPT-5.6-sol agents. Across the eight tasks with at least five GPT-5.6-sol runs, Qwen3.8-27B teams outscore five GPT-5.6-sol agents at the same price on Dabic Gravity Inversion, Graph Node Classification, Carleson Formalization and Active Learning (Figures~\ref{fig:cost} and~\ref{fig:cost-all}). GPT-5.6-sol stays ahead where its mean is higher, on Borden Source Inversion and Ad Placement Optimization, or its spread is wider, on KDA Chunk Forward. It also edges out Qwen3.8-27B on MoE Load Balance, 0.78 against 0.77. Pricing Qwen3.8-27B by its tokens instead of GPU time raises the count to 6 of 8.

A GPT-5.6-sol run is priced from its recorded token usage at list prices: \$4 per million input tokens, \$0.40 per million cached input tokens and \$20 per million output tokens, plus \$5 per million tokens written to cache. The CLI does not report cache writes, so every uncached input token is charged as a write, an upper bound. Qwen3.8-27B runs on our own GPUs, so a Qwen3.8-27B run is charged a flat share of the H200 engine that served it: a four-GPU engine at \$4.54 per GPU-hour serves twelve agents for 12 hours, \$18.16 per agent-run, and the kernel-task engines serve ten, \$21.79. Every Qwen3.8-27B agent therefore costs the same whatever it did with its time. The alternative is to price Qwen3.8-27B by tokens at list rates, \$0.25 per million input tokens, \$0.025 cached and \$2.50 output with the same cache-write charge, which comes to \$4 to \$29 per agent-run. Tokens price most Qwen3.8-27B runs below their GPU share, so the token basis favours Qwen3.8-27B against GPT-5.6-sol; it also favours independent Qwen3.8-27B agents against Qwen3.8-27B teams, whose members use 1.0 to 2.2 times the tokens of an independent agent, depending on the task. On the token basis the best Qwen3.8-27B configuration matches or beats GPT-5.6-sol on 6 of 8 tasks at the price of one GPT-5.6-sol agent and on 6 of 8 at the price of five, against 4 of 8 and 4 of 8 on the GPU basis.

Figure~\ref{fig:cost-all} repeats Figure~\ref{fig:cost} on the other six tasks with at least five GPT-5.6-sol runs. At the price of one GPT-5.6-sol agent, Qwen3.8-27B matches or beats it on MoE Load Balance and Active Learning and loses on Borden Source Inversion, Graph Node Classification, Ad Placement Optimization and KDA Chunk Forward. At five times the price it wins on Graph Node Classification and Active Learning and loses on Borden Source Inversion, Ad Placement Optimization, MoE Load Balance and KDA Chunk Forward, where one GPT-5.6-sol run reached 20.2. Two further tasks have a GPT-5.6-sol baseline, Kimi-Linear Decode and SwinMLP BF16, but only one run each, so they are not counted.

\section{Prompts}
\label{app:prompts}

Every agent receives one prompt at the start of its session. The harness assembles it from three pieces in a fixed order: the task's instruction file, a time-budget line, and, for a team member, a role note. A parallel agent receives the first two only. The same prompts go to every model; only the command-line harness differs. Nothing is pushed to an agent during a run. A team member learns about the forum from its role note alone and decides for itself when to read and when to post. If a session ends before the budget, the harness restarts it with a short resume prompt.

\paragraph{Task instruction.} Each task ships an instruction file that names the workspace, the deliverable, the submission command, its cooldown and the scoring rule. The box shows the file for Graph Node Classification. The other EdgeBench tasks follow the same skeleton with their own objective, deliverable path, cooldown and rules; the KernelBench, MLS-Bench and TopOpt tasks use their benchmarks' own instructions.

\begin{promptbox}{Task instruction (Graph Node Classification)}
# Iterative Evaluation Mode

You are working on a long-horizon research engineering task (from the EdgeBench
benchmark). Your work is scored by a hidden evaluator that you can invoke as
often as the cooldown allows -- submit early, submit often, and improve from the
feedback.

Your workspace is **`/home/workspace/045_graph_node_classification`** -- do all
task work there. **Read `README.md` and `solution/problem.md` in that workspace
first**: they hold the full task description, the dataset, the exact metric, and
the interface your solution must implement.

## The task

> Solve the graph node classification task described in `README.md` and
> `solution/problem.md`.

Your deliverable is a single file: **`solution/solution.py`**. The hidden judge
imports it and evaluates it on a withheld split -- so `solution/solution.py` must
be self-contained and runnable, and must implement the exact interface that
`solution/problem.md` specifies.

## Submitting & scoring

Run the evaluator with:

```bash
bash /app/submit.sh
```

This packages `solution/solution.py` and sends it to the hidden judge, which runs
`judge_eval.py` (imports your solution and evaluates it on the hidden split), then
prints your score and structured feedback. Notes:

- Only **`solution/solution.py`** is submitted (`submit_paths = "solution/solution.py"`).
  Make sure your best code is written there before every submission.
- **Score is on a 0-100 scale** (100 x the task metric); higher is better.
- **Submission cooldown: 120 seconds** between evaluations. Each evaluation may
  take a few minutes -- the judge trains/evaluates your model on **CPU (no GPU)**.
- The submit command waits for the result and prints it; if it ever times out
  waiting, the evaluation still finishes server-side and still counts.

## Scoring

- Your **best valid score across all submissions** is your final score. A
  submission that errors or is marked invalid does not count -- but it does not
  cost you points either.
- **Submit your latest work before your session ends** -- unsubmitted changes in
  the workspace do not count.

## Network environment

Treat this environment as having **no internet access** (the official benchmark
environment has none): do not use web search and do not fetch external resources.
Everything you need is already in the workspace.

## Rules

- Do not read or reference hidden test data, labels, or judge/scoring files.
- Do not try to read the judge's hidden split or hard-code its labels.
- Submit a self-contained `solution/solution.py` that reproduces your result.
\end{promptbox}

\paragraph{Time budget.} One line closes every prompt.

\begin{promptbox}{Time budget (appended to every prompt; 6 hours on Browser Exploit)}
---

## Time budget

Do not stop until 12 hours from now. You can check time with bash.
\end{promptbox}

\paragraph{Forum note.} Members of a team get one more section. It is the only mention of the forum an agent ever sees: nothing is pushed, and reading and posting are voluntary. In teams of 64 and 100 agents the index command shows the newest 100 titles and every unread post rather than the whole board.

\begin{promptbox}{Forum note (team of ten)}
## The shared forum (peer findings, on your terms)

You are one of 10 agents independently attacking this exact task -- each on your own node, each keeping your own history. A shared forum connects you: a persistent board of titled posts that any of you can write and any of you can read.

- **Post:** `python3 /app/comm.py post --title "one-line finding" --content "exact diff + measured score"` -- for anything multi-line, write the body to a file and pass `--content-file notes.md` instead of `--content` (shell-quoting a big body is how posts get mangled).
- **Scan:** `python3 /app/comm.py list` -- the title index, with NEW markers for posts you haven't seen yet.
- **Read:** `python3 /app/comm.py read p00042` -- one full post.

Norms that make the forum worth having:

- Post when you hit a clear WIN or a clear FAILURE -- negative results save your teammates GPU-hours; "tried X, score got worse" is as valuable as a win.
- Include the exact config diff and the measured `score`. A post a teammate can act on beats a claim of progress.
- Run `python3 /app/comm.py list` before starting each new experiment -- a teammate may already have tried your idea.
- Reading a full post costs tokens; titles are cheap. Scan the list first, then `read` only the posts that earn it.

Posting and reading are entirely your choice -- nobody will post for you, and nobody will read for you.
\end{promptbox}

\paragraph{Resume prompt.} When a session ends before the budget, for example at the harness's turn cap, the harness restarts it with the task's resume prompt.

\begin{promptbox}{Resume prompt (Graph Node Classification)}
Continue working on the EdgeBench task. Your workspace is
`/home/workspace/045_graph_node_classification` -- re-read `/app/instruction.md`,
`README.md`, and `solution/problem.md` there if you need to reorient.

Keep improving your **`solution/solution.py`** and, after every meaningful change,
submit it with `bash /app/submit.sh` (120-second cooldown between submissions).
The judge imports your `solution/solution.py` and evaluates it on a hidden split;
your best valid score across all submissions is your final score, so always
submit your latest working state before you run out of time. Do not stop early.
\end{promptbox}

\paragraph{Orchestrated teams.} In the orchestrator ablation of Section~\ref{sec:ablate-harness}, the forum note is replaced by one of two role notes. The orchestrator plans and assigns; the workers wait, execute and report. The workers never see each other.

\begin{promptbox}{Orchestrator note (one orchestrator, nine workers)}
## Your role: ORCHESTRATOR of a 10-agent team

You are `agent_0`, the orchestrator. 9 worker agents -- agent_1, agent_2, agent_3, agent_4, agent_5, agent_6, agent_7, agent_8, agent_9 -- each run in their own identical copy of this environment (same task, same tools, same submission command, their own workspace and their own scores). They do not talk to each other, and they will not do anything until you assign them something. You are the central planner: read the task, split it into pieces, hand each worker a piece, wait for the results, and re-plan from what comes back.

The channel is `/app/orch.py` (only you and the workers see it; every message persists on the bus):

- `python3 /app/orch.py assign agent_1 --title "one-line goal" --content-file task.md` -- give a worker its next task (`--content "..."` for short bodies; `all` instead of an id sends the same assignment to every worker). A new assignment supersedes the worker's current one.
- `python3 /app/orch.py wait` -- block until something needs you: a worker reports back, or a worker has been left with nothing to do (it returns early and names them; an idle worker sends no message of its own, so this is the only way you hear about one). Returns after 4 minutes if neither happened -- call it again. Prints each report in full, then every worker's state. Run it as a normal foreground command and let it block (give your shell tool a timeout longer than that); do not background it -- nothing will notify you, and a turn that ends while you are waiting is simply restarted.
- `python3 /app/orch.py status` -- every worker's state: unassigned / working (since when, on what) / idle (last report).
- `python3 /app/orch.py note agent_1 --title "..." --content "..."` -- a message that is not a new assignment (an answer, a correction, a heads-up).
- `python3 /app/orch.py history [--full]` -- the whole conversation so far. Also how you recover your bearings after a context reset or a sandbox restart: nothing is lost, it is all on the bus.

The loop you run until the budget ends: give every worker an assignment -> `wait` -> read the report -> decide what that worker should do next (continue, pivot, build on what another worker found, verify, finalize) -> `assign` -> `wait` again. Assignments should be concrete enough to act on: what to try, what to measure, what to send back. Reports carry measured scores and can carry full code (bodies up to 64,000 characters), so you can hand a strong result from one worker to another verbatim.

What counts is each worker's own best valid submission in its own workspace; the team's result is the best worker. Your own workspace is not graded -- your leverage is the plan, the division of labor, and what you do with the results. Keep every worker busy on a distinct, worthwhile piece, react as soon as a report lands, and leave time at the end for each worker to lock in its best submission.

## You are the mastermind: the team is a portfolio of bets

Think of the team as a portfolio rather than a queue. Each worker is a bet on a different way this task might be won, and you are spending a fixed budget of worker-hours across those bets. Two workers grinding the same idea is one bet at double the price, so it is usually worth opening on genuinely different hypotheses -- a different algorithm or formulation, a different representation, a different place to spend the compute -- and making each one concrete enough that it can be proved wrong quickly.

Every report is information about the whole portfolio, not only about the bet that reported. From one you might back the line that is compounding, give a stalled one a sharpened retry, drop one that is dead, or hand a strong result to another worker verbatim so that a single good idea gets developed along several branches. How you spread the budget across the bets is yours to judge, and it does not have to stay even or stay fixed.
\end{promptbox}

\begin{promptbox}{Worker note (one of nine workers)}
## Your role: WORKER in a 10-agent team

You are `agent_1`, one of 9 workers directed by an orchestrator agent (`agent_0`). The orchestrator has read the same task and decides what you work on; you do not choose your own experiments. You do not see the other workers and they do not see you -- the orchestrator is your only channel.

`/app/orch.py` is that channel:

- `python3 /app/orch.py wait` -- block until your next assignment arrives (returns after 4 minutes if nothing came -- call it again). Run it as a normal foreground command and let it block (give your shell tool a timeout longer than that); do not background it -- nothing will notify you, and a turn that ends while you are waiting is simply restarted.
- `python3 /app/orch.py report --title "one-line result" --content-file report.md` -- send your result back when an assignment is done (`--content "..."` for short bodies).
- `python3 /app/orch.py inbox` -- check for new messages without blocking (an assignment can arrive while you are still working on the previous one; the newest one is the one to follow).
- `python3 /app/orch.py note --title "..." --content "..."` -- a question or a progress update that is not a final report.

Your loop: orient yourself in the task materials first, then `wait`. When an assignment arrives, do exactly that piece of work in your workspace -- running the task's normal submission command so your work is scored -- then `report` what you did, what you measured (scores, numbers, what failed) and anything the orchestrator needs in order to plan the next step (full code fits: bodies go up to 64,000 characters). Then `wait` again. Waiting is part of the job: do not start work of your own choosing while idle, and do not stop -- an assignment can arrive at any time until the budget ends.

Your own best valid submission is your score; the team's result is the best worker. Always keep your best solution submitted, especially near the end of the budget.
\end{promptbox}

\end{document}